\documentclass[10pt,journal,compsoc]{IEEEtran}
\ifCLASSOPTIONcompsoc
  \usepackage[nocompress]{cite}
\else
  \usepackage{cite}
\fi

\ifCLASSINFOpdf
\else
\fi

\usepackage{amsmath}
\usepackage{amssymb}
\usepackage{amsfonts}
\usepackage{mathtools}
\usepackage{bbm}
\usepackage{dsfont}
\usepackage{nicefrac}

\usepackage{graphicx}
\usepackage{epsfig}
\usepackage[export]{adjustbox}
\usepackage{xcolor}

\usepackage[utf8]{inputenc}
\usepackage[T1]{fontenc}
\usepackage{microtype}
\usepackage{anyfontsize}

\usepackage{array}
\usepackage{booktabs}
\usepackage{multirow}
\usepackage{makecell}
\usepackage{tabularx}
\usepackage{float}
\usepackage{wrapfig}
\usepackage{capt-of}

\usepackage[caption=false,font=normalsize,labelfont=sf,textfont=sf]{subfig}

\usepackage{stfloats}

\usepackage{enumitem}
\usepackage{xspace}
\usepackage{pifont}
\usepackage{textcomp}
\usepackage{url}
\usepackage{verbatim}

\usepackage{cite}

\usepackage{algorithm}
\usepackage{algorithmicx}
\usepackage{algpseudocode}

\usepackage{pifont}
\usepackage{xcolor}
\usepackage{algorithmicx}
\usepackage{algorithm}
\usepackage{algpseudocode}

\newcommand{\cmark}{\textcolor{green!50!black}{\checkmark}}   %
\newcommand{\xmark}{\textcolor{red!70!black}{\ding{55}}}      %
\usepackage{xspace}
\newcommand{\METHOD}{{QuPAINT\xspace}}

\usepackage{hyperref}
\usepackage{cleveref}

\begin{document}

\title{QuPAINT: Physics-Aware Multimodal Reasoning for Quantum Material Characterization\textsuperscript{*}}

\author{
Sankalp~Pandey,
Xuan-Bac~Nguyen,
Hoang-Quan~Nguyen,
Tim~Faltermeier,\\
Nicholas~Borys,
Hugh~Churchill,
and~Khoa~Luu
\\[-0.15em]
{\normalsize\url{https://uark-cviu.github.io/projects/qupaint/}}

\thanks{
Sankalp Pandey, Xuan-Bac Nguyen, Hoang-Quan Nguyen, and Khoa Luu
are with Quantum AI Lab in the Department of Electrical Engineering \& Computer Science,
University of Arkansas, Fayetteville, AR 72701, USA.
E-mail: \{sankalpp, xnguyen, hn016, khoaluu\}@uark.edu.
}
\thanks{
Hugh Churchill is with the Department of Physics,
University of Arkansas, Fayetteville, AR 72701, USA.
E-mail: hchurch@uark.edu.
}
\thanks{
Tim Faltermeier and Nicholas Borys are with the University of Utah,
Salt Lake City, UT, USA.
E-mail: timfaltermeier@montana.edu, nicholas.borys@utah.edu.
}
\thanks{
The authors are affiliated with the MonArk NSF Quantum Foundry.
}

\thanks{\raisebox{0.35ex}[0pt][0pt]{\normalsize *}\,A preliminary version of this work was accepted to the Findings Track of the IEEE/CVF Conference on Computer Vision and Pattern Recognition (CVPR), 2026~\cite{nguyen2026qupaint}. This manuscript substantially extends the conference version.
}
}

\IEEEtitleabstractindextext{%
\begin{abstract}
Characterizing two-dimensional (2D) quantum materials by optical microscopy requires localizing exfoliated flakes and determining their layer thickness from subtle optical contrast and interference color to select suitable flakes for device fabrication. However, models face synthetic-to-real domain shifts and variation across materials, substrates, laboratories, and imaging conditions. We present QuPAINT, a physics-aware multimodal framework for transferable quantum flake characterization. The Synthetic Materials Framework (Synthia) generates diverse synthetic microscopy images while preserving layer-dependent optical behavior. Using these images, we construct QMat-Instruct, a multimodal instruction dataset with image-specific reasoning traces generated from verified annotations and constrained to observable optical cues. QuPAINT integrates these signals through Physics-Informed Attention (PIA), which injects substrate-relative optical priors into the visual representation to support grounded multimodal reasoning. For evaluation, we introduce QF-Bench, to our knowledge, the largest real-world benchmark for this problem, spanning diverse microscopy and substrate conditions. Using its verified annotations, we study counting, visual grounding, reasoning quality, confidence calibration, and transfer to an unseen material. QuPAINT-8B substantially outperforms prior methods and establishes state-of-the-art performance for both general and monolayer flake detection. Additional experiments show that image-grounded supervision improves strict spatial grounding and confidence calibration while preserving robust general flake detection on the unseen material.
\end{abstract}

}

\maketitle

\begin{figure*}[!t]
    \centering
    \includegraphics[width=0.95\linewidth]{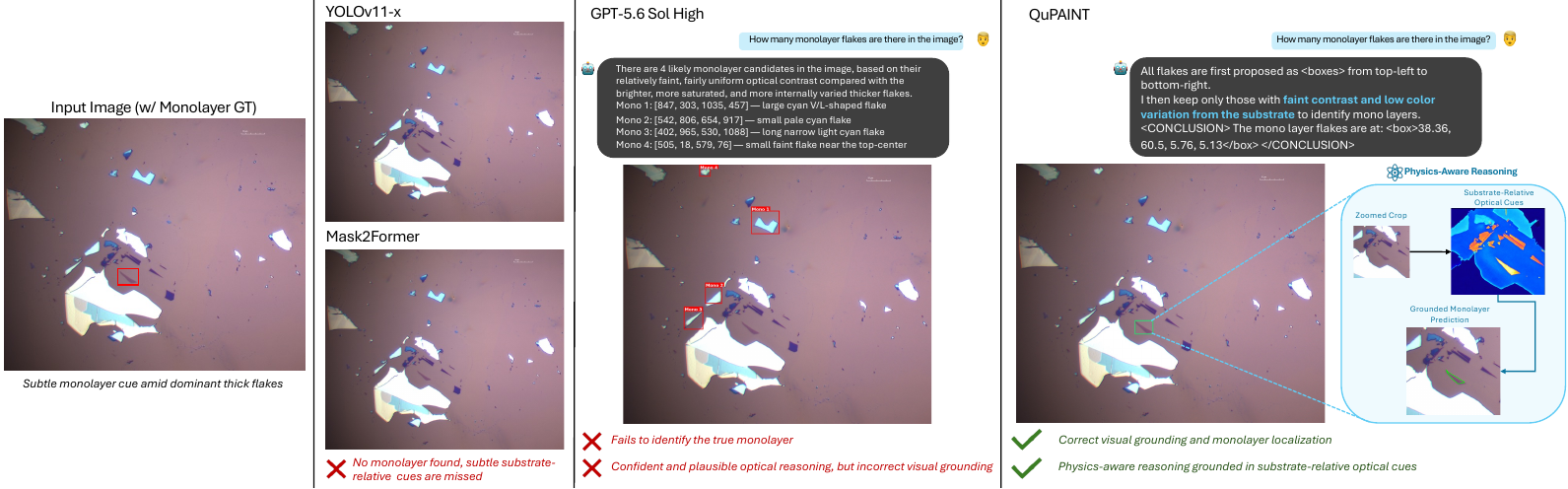}
   \caption{\textbf{Failure modes of conventional detectors and general-purpose commercial-grade vision--language models for monolayer identification.}
Both the CNN-based YOLOv11-x detector and the Transformer-based Mask2Former detector fail to confidently locate any candidate monolayers, while GPT-5.6 Sol High provides plausible optical reasoning but incorrect visual grounding. \METHOD{} instead grounds its prediction in substrate-relative optical cues and correctly localizes the monolayer. \textbf{(Best viewed in color.)}}
    \label{fig:hero}
\end{figure*}

\section{Introduction}
\label{sec:intro}

Atomically thin two-dimensional (2D) quantum materials such as graphene, $\text{MoS}_2$, $\text{WTe}_2$, h-BN, and related systems have emerged as a foundation for a new generation of electronic, photonic, and quantum devices due to properties such as quantum confinement, spin-valley coupling, and layer-dependent band topology~\cite{ISLAM2025100161}. These properties depend on the thickness, or number of layers, of the material flakes. 

\noindent
\textbf{Why is this Problem Challenging?} In optical microscopy, however, the layer count of a flake is not directly observed. Locating a flake is only the first step. The main challenge is determining its thickness from minute changes in reflected color and intensity produced by thin-film interference. Adjacent layer categories can have nearly identical shape and texture and differ only through weak optical contrast relative to the surrounding substrate, while the observed appearance of the same layer can change substantially with material type, oxide thickness, illumination, camera response, focus, and white balance.
Thus, building devices with exfoliated flakes relies on being able to answer a deceptively simple question: \emph{how many layers is this flake?} \Cref{fig:hero} illustrates how this remains difficult for conventional detectors and general-purpose vision--language models, and how \METHOD{} addresses the problem through physics-aware, visually grounded reasoning.

\begin{figure}[!t]
    \centering
    \includegraphics[width=\linewidth]{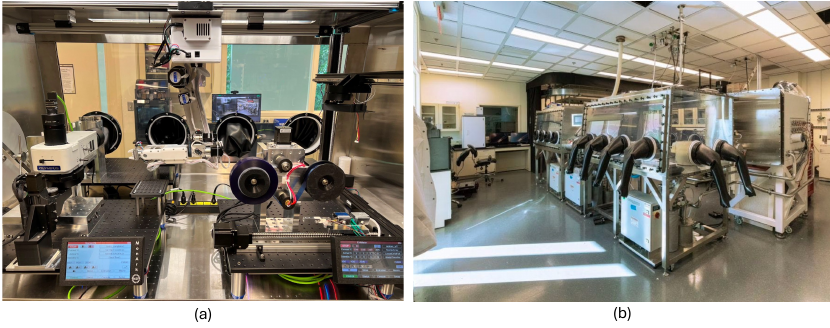}
    \caption{\textbf{Real experimental infrastructure for 2D quantum material characterization.}
    (a) Automated optical microscopy platform for imaging and characterizing exfoliated 2D material samples.
    (b) Glovebox-based experimental infrastructure for material preparation and handling under controlled conditions.
    Together, these systems illustrate the physical experimental workflow that requires scalable, automated flake identification and layer characterization. Images courtesy of the MonArk NSF Quantum Foundry 
    \cite{monarkfoundry}.
    \textbf{(Best viewed in color.)}}
\label{fig:experimental_infrastructure}
\end{figure}

\begin{figure*}[!t]
\centering
\includegraphics[width=1\linewidth]{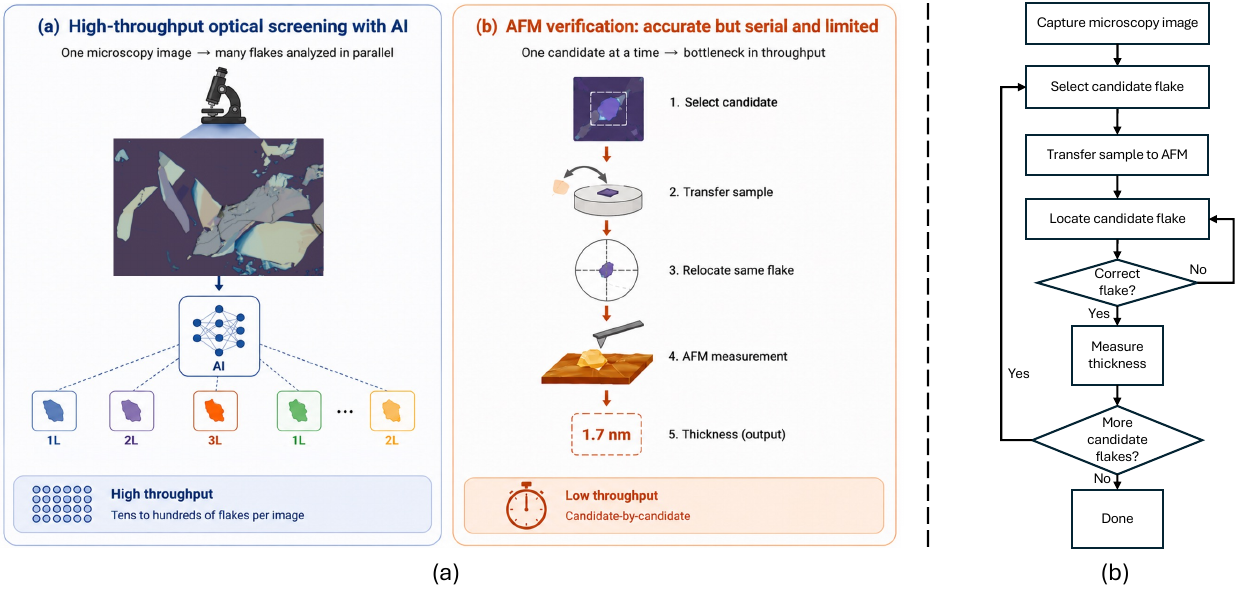}
\caption{\textbf{Overview of the real-data characterization pipeline and challenges in 2D material analysis.}
(a) Comparison between optical microscopy with AI-assisted analysis and AFM-based manual characterization. Optical microscopy enables rapid, scalable thickness prediction across many flakes, whereas AFM requires manual acquisition and measurement for individual flakes.
(b) The conventional AFM characterization workflow requires repeated candidate selection, sample transfer, flake relocation, verification, and thickness measurement, resulting in substantial manual effort.
\textbf{(Best viewed in color.)} 
}
\label{fig:dataset_challenging}
\end{figure*}

In practice, quantum materials researchers must repeatedly search optical microscopy images to find suitable flakes and estimate their thickness from subtle variations in optical contrast and interference color. This process is time-consuming and can be inconsistent across materials, substrates, laboratories, and imaging conditions. The experimental infrastructure supporting this workflow, including automated optical microscopy and controlled material preparation, is shown in \Cref{fig:experimental_infrastructure}. Moreover, reliable thickness confirmation often requires transferring the sample to an Atomic Force Microscope (AFM). Although AFM provides accurate layer measurements, it adds a characterization step that further limits device-fabrication throughput and scalability, as illustrated in \Cref{fig:dataset_challenging}.

\subsection{Extension of the Conference Version}
This journal article extends our conference work~\cite{nguyen2026qupaint} in both the supervision used to train the model and the evaluation used to study its behavior. In the conference version, we generated reasoning responses in QMat-Instruct using predefined templates. Although these templates provided structured physics-aware supervision, samples with the same layer category could receive similar explanations regardless of the specific optical properties visible in the image.

In this work, we generate an image-specific reasoning trace for each training sample using a vision--language model provided with the verified ground-truth annotations. The generated response is constrained to evidence visible in the microscopy image, including relative contrast, color, boundaries, and appearance with respect to the surrounding substrate. It is also instructed not to rely on measurements that cannot be obtained from the input image, such as Raman spectroscopy, photoluminescence, AFM, or other external characterization techniques. This produces more diverse reasoning supervision while preserving the verified flake locations and layer labels.

We further extend the evaluation beyond conventional detection performance. In addition to measuring whether the model correctly detects and classifies each flake, we study instruction following, spatial grounding, reasoning quality, and transfer across material systems. We also evaluate whether the model can generalize when one material is excluded from training. These experiments provide a more complete analysis of whether the model has learned transferable optical relationships rather than only patterns specific to the training dataset.

\subsection{Why Conventional Vision Models Struggle?}
\begin{figure}[!htbp]
\centering
\includegraphics[width=0.95\linewidth]{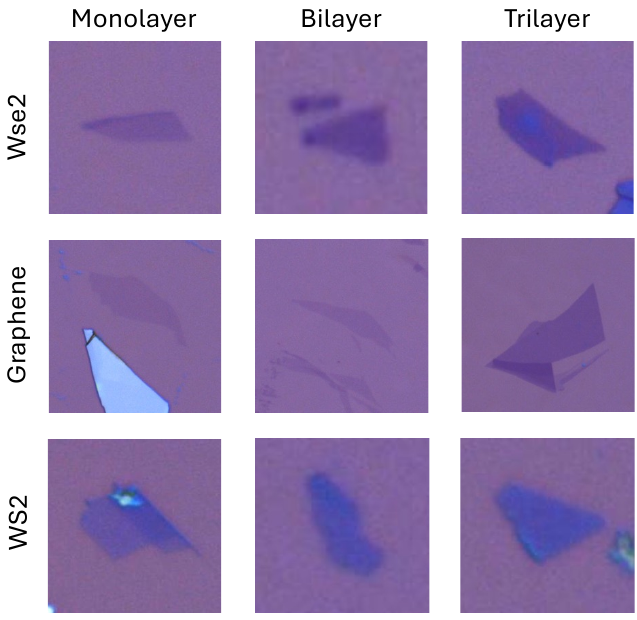}
\caption{Optical microscopy images of monolayer, bilayer, and trilayer flakes across different 2D materials exhibit only subtle visual differences, highlighting the difficulty of determining layer thickness directly from appearance. \textbf{(Best viewed in color.)}
}
\label{fig:visual_difficulty}
\end{figure}

Modern vision models, including CNN-based~\cite{he2017mask,redmon2016you} and Transformer-based~\cite{robinson2025rfdetrneuralarchitecturesearch,ravi2024sam,peng2024dfineredefineregressiontask,Zhou_2025_ICCV,assran2025vjepa2selfsupervisedvideo,siméoni2025dinov3} approaches, are highly effective when object categories are supported by stable visual cues such as shape, texture, color, and spatial structure. In quantum flake characterization, however, these cues are often insufficient for distinguishing layer categories. A model may localize the flake correctly while still failing to infer its thickness, because mono-layer, bi-layer, and tri-layer regions can share nearly identical geometry and differ primarily through weak optical changes, as illustrated in \Cref{fig:visual_difficulty}.

A second difficulty is that the discriminative cues are not stable across acquisition settings.
Changes in illumination, camera response, substrate and oxide thickness, focus, and white balance can alter the observed appearance more strongly than the difference between neighboring layer categories. Consequently, a model trained on absolute appearance can learn acquisition-specific correlations rather than the relative optical relationship between the flake and its surrounding substrate.

The learning problem is further complicated by limited and highly imbalanced supervision. Verified mono- and few-layer flakes are rare, while thick flakes dominate typical microscopy datasets. This makes high-capacity models prone to overfitting to the dominant classes or laboratory-specific appearance statistics, rather than learning the subtle cues required for layer identification.

Finally, the desired output is not only a bounding box or semantic category. The model must infer a physical quantity, the number of layers, from weak optical evidence and provide a prediction that can be inspected and verified. These limitations motivate a formulation that incorporates physical priors and relative optical reasoning rather than relying only on conventional visual pattern recognition.

\subsection{Limitations of Prior Work}
Existing domain-specific systems~\cite{masubuchi2020deep,uslu2024open,uslu2025maskterial} demonstrate the potential of automated flake characterization. However, these methods are developed and evaluated using a limited set of materials, substrates, laboratories, or imaging conditions. In real experimental settings, changes in microscope illumination, camera sensors, magnification, substrate configuration, and material type can substantially alter the optical appearance of the flakes. As a result, models trained using one acquisition pipeline generally do not transfer reliably to another laboratory or to unseen material systems.

The field also lacks an established and standardized protocol for measuring transfer across these domains. Existing methods are evaluated using different datasets, materials, and experimental configurations, making direct comparison difficult. It is therefore unclear whether their performance reflects transferable relationships between optical appearance and flake thickness or adaptation to a specific acquisition pipeline. These limitations motivate a framework that can learn from scarce, imbalanced supervision, incorporate physical priors, and generalize across laboratories, materials, substrates, and imaging conditions.

\subsection{Physics-Aware Reasoning}
We formulate quantum flake characterization as a physics-aware multimodal reasoning task rather than only an object detection problem. Given an optical microscopy image, a Multimodal Large Language Model (MLLM) learns to compare the visible optical properties of each flake with the surrounding substrate and produce an interpretable response containing the number of flakes, their locations, their predicted layer categories, and the rationale behind each prediction.

\begin{figure*}[!t]
    \centering
    \includegraphics[width=0.95\linewidth]{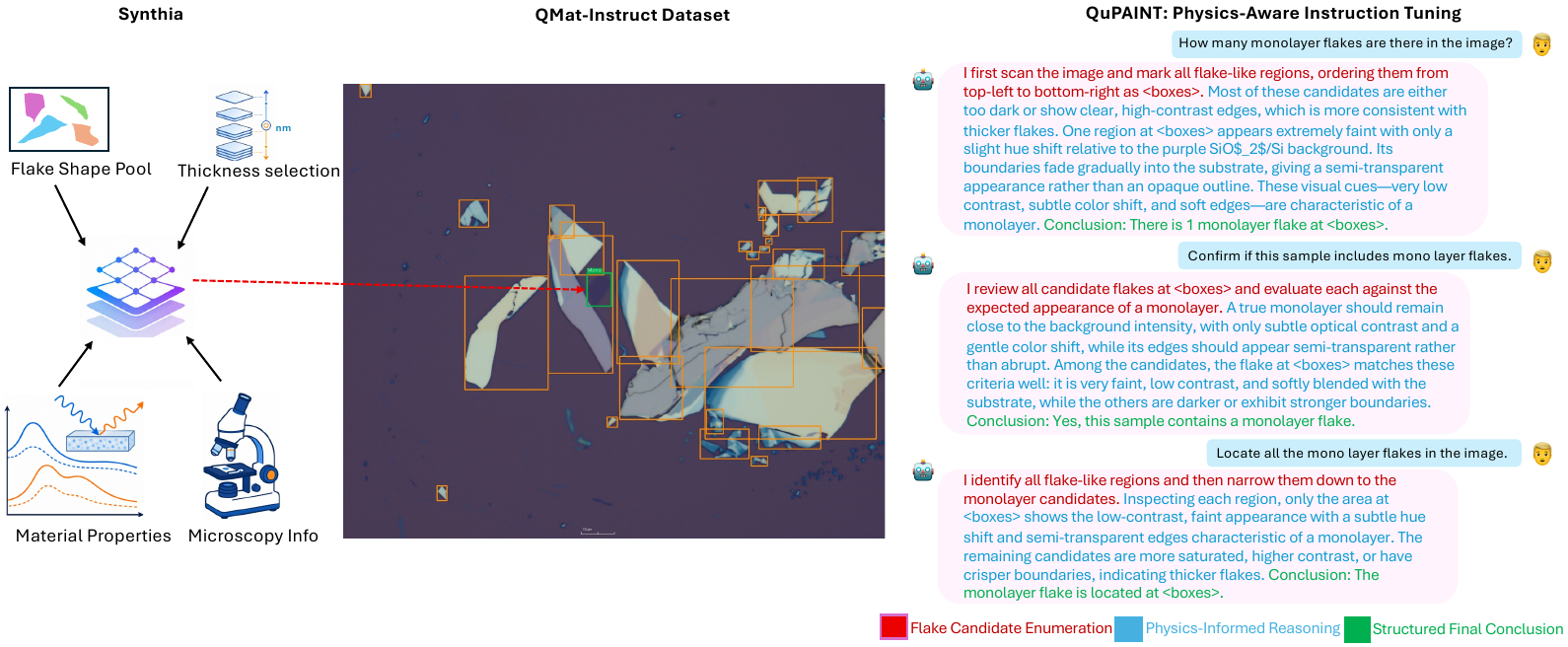}
    \caption{
        \textbf{Overview of the proposed \METHOD{} framework.}
        \textbf{Synthia} generates physics-grounded synthetic microscopy images with realistic flake appearances, while \textbf{QMat-Instruct} provides multimodal, physics-aware instruction supervision. These components support \textbf{\METHOD}, which combines physics-informed visual representations with multimodal reasoning for grounded quantum material characterization.
        \textbf{(Best viewed in color.)}
    }
    \label{fig:qupaint_framework}
\end{figure*}

This formulation allows the model to use relative optical cues rather than relying only on absolute color or contrast values. In particular, the model can explicitly reason about whether a candidate region is brighter, darker, or differently colored than its surrounding substrate and relate these differences to the predicted layer category. The resulting response provides both the prediction and the evidence supporting it. \Cref{fig:qupaint_framework} summarizes the overall framework, including the physics-based data generation, multimodal instruction supervision, and physics-informed reasoning components used by \METHOD{}.

\subsection{Contributions of this Work}
In summary, this work has four main contributions. First, we introduce \textbf{Synthetic Materials Framework (Synthia)}, a physics-based data generation framework that renders realistic quantum material flakes using thin-film interference modeling, CIE color representations, and substrate-specific color calibration. The generated flakes are composited onto real microscopy images while avoiding overlap with existing flakes. Synthia increases the number and diversity of rare mono- and few-layer examples that are difficult to collect and label through real experiments.

Second, we construct \textbf{QMat-Instruct}, a large-scale multimodal instruction dataset for quantum material characterization. We replace fixed reasoning templates with image-specific reasoning traces generated from verified annotations. The generated responses are constrained to optical evidence visible in the image and describe each flake using relative differences with respect to the surrounding substrate. This keeps the reasoning supervision applicable across different materials and substrate configurations without directly providing material identity or oxide thickness.

Third, we propose \textbf{Physics-Aware Instruction Tuning (\METHOD)}, a physics-aware multimodal framework containing a Physics-Informed Attention (PIA) module. PIA uses the relative LAB contrast between each image region and its surrounding substrate to emphasize physically meaningful visual cues. This allows \METHOD{} to focus on faint mono- and few-layer flakes rather than simply prioritizing regions with the strongest absolute contrast. 

Finally, we establish \textbf{QF-Bench}, a multi-material benchmark for optical microscopy-based characterization of 2D material flakes. QF-Bench contains 280,526 annotated flakes across eight material systems and multiple experimental settings. To the best of our knowledge, it is the largest real-world benchmark for this problem. Using QF-Bench, we evaluate conventional flake detection together with instruction following, spatial grounding, reasoning behavior, and transfer to a material held out from training.
By bringing diverse materials and acquisition settings into a single benchmark, QF-Bench provides the community with a common testbed for reproducible comparison across methods and for measuring whether performance transfers beyond the conditions observed during training. This addresses a major limitation of existing evaluation protocols and provides a stronger foundation for developing generalizable automated 2D material characterization systems for real-world use.

\section{Related Work}

\subsection{Automated 2D Material Characterization}
The promise of 2D materials for quantum technologies and algorithms \cite{nguyen2024two,nguyen2024quantum,nguyen2024qclusformer,holliday2024hybrid,nguyen2024hierarchical,nguyen2025diffusion,nguyen2025qmoe,holliday2025quadro,pandey2025cliff,holliday2026advanced,nguyen2026quacod} has driven sustained interest in identifying and characterizing exfoliated flakes with automated inspection \cite{dendukuri2020definingquantumneuralnetworks, Ouaj_2024, Ouaj_2023, steiner2025current, mckenzie2024fabrication, jackering2025super, PALINKAS2024119608, Hattori_2024, wegerhoff2025coherent, crimmann2025high, uslu2025maskterial, courtney2025automated}.
The problem was first cast as dense prediction, with UNet variants \cite{ronneberger2015u, han2020deep} and Mask R-CNN detectors \cite{he2017mask, masubuchi2020deep} operating directly on optical micrographs, and was later refined through attention with soft labels \cite{nguyen2024two}, explicit handling of missed flakes \cite{luu2024automatically}, and extension to compositionally heterogeneous samples \cite{he2025rapid}.
Classical alternatives such as Gaussian mixture models offer interpretability \cite{uslu2024open} but degrade under the illumination and noise variation of real acquisition.
Scaling this pipeline has taken two forms: instrumentation, where automated microscopes survey large substrate areas to support statistical analysis \cite{crimmann2025high}, and representation, where 2D-material foundation models \cite{uslu2025maskterial} and general segmentation backbones \cite{kirillov2023segment, ravi2024sam} supply transferable priors.
Increasingly, characterization is treated not as a standalone task but as one stage of a closed loop spanning exfoliation, sample handling, and downstream device assembly \cite{mckenzie2024fabrication, Ouaj_2024, Ouaj_2023, jackering2025super, courtney2025automated, atomic2025autonomous, aila2025autonomous}.
Two obstacles limit how far this progress carries.
Accuracy achieved on one material rarely survives the move to another, a failure that continual-learning formulations mitigate by preserving earlier materials while accommodating new ones \cite{pandey2025cliff}, and models fit to simulated flakes lose calibration on measured data unless adaptation is anchored in the optics that produce flake contrast \cite{phiadapt2025physics}.

\subsection{Multimodal Large Language Models (MLLMs)}
A parallel line of work equips language models to reason over perceptual input.
Contrastive pretraining first established shared image-text representations \cite{radford2021learning, jia2021scaling}, after which instruction tuning coupled frozen visual encoders to pretrained LLMs and opened the door to free-form visual dialogue \cite{li2023blip, liu2023visual, dai2023instructblip}.
Subsequent efforts have targeted deliberate reasoning \cite{xu2025llava}, cooperation among specialized agents \cite{dong2025insight}, downstream task alignment \cite{yan2025task, tan2025beyond}, and robustness when the test distribution departs from training \cite{xu2025rdmldg}, while proprietary and open systems at scale have broadened coverage from static images to video \cite{achiam2023gpt, bai2023qwen, chen2024internvl}.
Handling video has in turn motivated dedicated treatments of time, including state space models for long motion sequences \cite{zhang2024motion}, event-level temporal grounding \cite{qian2024momentor}, and training-free frame selection for long-form inputs \cite{liu2025bolt}.
Transferring these capabilities to science has proven harder.
Materials-oriented systems have begun to pair structural representations with language for property prediction \cite{matterchat2026multimodal}, and a set of benchmarks now probes multimodal reasoning over authentic characterization data \cite{matcha2025multimodal, matqna2025multimodal, macbench2025probing, see2026science}.
Their collective picture is that factual recall and routine extraction are largely solved, whereas subtle perceptual discrimination and inference chained through physical constraints are not \cite{matcha2025multimodal, macbench2025probing}.
This is precisely the regime that layer counting occupies, where the evidence is a few percent of contrast rather than an object category, and where purely vision-centric models \textit{struggle to generalize across domains}.
We therefore extend our conference framework \cite{nguyen2026qupaint} with a physics-informed multimodal design that ties observed appearance to the optical properties responsible for it.

\section{Challenges and Limitations of Existing Quantum Materials Datasets}

\subsection{Challenges in Data Collection}
Two-dimensional (2D) quantum materials, including graphene and transition-metal dichalcogenides (TMDs), are commonly produced through mechanical exfoliation. In this process, a bulk crystal is repeatedly cleaved using adhesive tape until atomically thin flakes are obtained and transferred onto a substrate, such as SiO$_2$/Si. Although mechanical exfoliation is simple and inexpensive, the process is inherently random. The resulting flakes can have different shapes, sizes, and thicknesses, with no direct control over where flakes with a particular number of layers will appear.

\begin{table}[!t]
\centering
\caption{Comparison of dataset synthesis techniques between MaskTerial~\cite{uslu2025maskterial} and Synthia.}
\label{tab:synthesis_comparison}
\resizebox{\columnwidth}{!}{
\begin{tabular}{lcc}
\toprule
\textbf{Technique Aspect} 
& \textbf{MaskTerial~\cite{uslu2025maskterial}} 
& \textbf{Synthia (Ours)} \\
\midrule
Flake shapes from multi-material sources 
& \cmark & \cmark \\
Physics-based optical simulation (TMM) 
& \cmark & \cmark \\
Personalized color calibration 
& \xmark & \cmark \\
Using CIE standard 
& \xmark & \cmark \\
Avoid overlapping with existing flakes 
& \xmark & \cmark \\
\bottomrule
\end{tabular}
}
\end{table}

Producing a large number of random flakes is relatively simple, as a single exfoliation attempt can generate thousands of flakes. However, constructing a dataset containing flakes with specific layer numbers, such as monolayer, bilayer, or few-layer regions, is much more difficult. Because the exfoliation process does not provide direct control over flake thickness, researchers must search among a much larger number of thick flakes, artifacts, and other unsuitable regions.

This flake-hunting and labeling process is time-consuming. Optical microscopy images can be collected quickly, but they do not directly measure layer thickness. Instead, researchers estimate thickness from optical contrast and interference color, both of which can change with microscope illumination, camera response, substrate configuration, and other imaging conditions. Reliable thickness confirmation therefore commonly requires Atomic Force Microscopy (AFM). Each candidate flake must be relocated under the AFM, measured at the nanometer scale, and then matched back to its location in the original optical microscopy image. This alignment between optical and AFM measurements requires substantial manual effort and limits data-collection throughput.

As a result, constructing quantum flake datasets for AI training is both labor-intensive and difficult to scale. The combination of random flake generation, uncertain thickness estimation from optical images, and slow AFM-based verification limits the amount of accurately labeled data that can be collected. \Cref{fig:dataset_challenging} provides an overview of this data-collection process.

\subsection{Limitations of Existing Synthetic Datasets}
MaskTerial~\cite{uslu2025maskterial} introduced physics-based rendering for generating large-scale synthetic datasets of 2D material flakes. However, its synthesis pipeline has several limitations that can increase the difference between synthetic and real microscopy images. First, its optical simulation does not include image-specific color calibration, which can cause the simulated flake colors to differ from those observed under a particular microscope. Second, it does not explicitly use the CIE color standard when converting the simulated optical response into image colors. This can limit how well the generated data transfers across microscopes with different illumination and camera settings.

In addition, the synthetic flakes are placed without explicitly accounting for flakes already present in the background image. This can produce unrealistic overlap or physically inconsistent stacking between synthetic and existing flakes. \Cref{tab:synthesis_comparison} summarizes the differences between the previous synthesis method and our proposed framework.

\section{Synthetic Materials Framework (Synthia)}

In this section, we introduce \textbf{Synthia}, a physics-based synthetic data generation framework for 2D quantum materials. Synthia generates realistic microscopy images containing flakes with controlled material types, layer numbers, locations, and optical properties. It combines a multilayer optical model with color calibration and substrate-aware flake placement to reduce the difference between synthetic and real microscopy images. The images and verified annotations generated by Synthia are then used to construct \textbf{QMat-Instruct}, a multimodal instruction dataset for training MLLMs to identify, localize, and reason about quantum flakes.

\subsection{Multilayer Optical Model}
\label{sec:tmm}

To generate realistic images of 2D material flakes on common substrates, we model their optical appearance using the \textit{Transfer Matrix Method} (TMM)~\cite{byrnes2020multilayeropticalcalculations}. TMM models the thin-film interference produced by the material, oxide layer, and substrate. This allows us to compute the wavelength-dependent reflectance and optical contrast for flakes with different thicknesses.

\noindent\textbf{Optical Model.}
Consider a multilayer stack containing $L$ thin layers, indexed by $l=1,2,\dots,L$. The incident medium and substrate are represented by layers $0$ and $L{+}1$, respectively, and are assumed to be semi-infinite. Each layer $l$ is defined by its complex refractive index $n_l(\lambda)+ik_l(\lambda)$ and physical thickness $d_l$.

When light with wavelength $\lambda$ enters the stack at normal incidence, it undergoes multiple reflections and transmissions at the interfaces between the layers. At the interface between layers $l$ and $l{+}1$, the reflection and transmission coefficients are determined by the Fresnel equations as 
in \Cref{eq:fresnel}:
\begin{equation}
\label{eq:fresnel}
r_{l,l+1} = \frac{n_l - n_{l+1}}{n_l + n_{l+1}},
\quad\quad
t_{l,l+1} = \frac{2n_l}{n_l + n_{l+1}}.
\end{equation}

Propagation through each layer introduces a phase delay
$\delta_l = \frac{2\pi n_l d_l}{\lambda}$,
which is represented using a propagation matrix $P_l$. The total electric field across the multilayer stack can then be computed using the transfer-matrix formulation in \Cref{eq:tmm_matrix}:

\begin{equation}
\begin{pmatrix}
v \\[3pt] u
\end{pmatrix}
= M_{0,1}
\left( \prod_{l=1}^{L} P_l \, M_{l,l+1} \right)
\begin{pmatrix}
u_0 \\[3pt] 0
\end{pmatrix},
\label{eq:tmm_matrix}
\end{equation}

\noindent where $u_0=1$ is the incident field amplitude, $v$ is the reflected field amplitude, and $u$ is the transmitted field amplitude in the substrate. The wavelength-dependent reflectance is then computed as
$R(\lambda)=\left|v/u_0\right|^2$.

\noindent\textbf{Reflectance Computation for Synthetic Flakes.}
When we synthesize flakes, each flake is represented as a multilayer stack consisting of \textbf{(1)} air as the incident medium, \textbf{(2)} a 2D material layer with thickness $d_\text{flake}$ determined by its number of layers, \textbf{(3)} an oxide layer, such as SiO$_2$, with thickness $d_\text{SiO2}$, and \textbf{(4)} a semi-infinite silicon substrate.

We sample wavelengths $\lambda\in[400,700]$~nm at $D$ discrete intervals and compute $R(\lambda)$ using \Cref{eq:tmm_matrix}. Repeating this process across different materials and layer thicknesses produces the reflectance spectra used to determine the appearance of the synthetic flakes.

\noindent\textbf{Color-Space Conversion.}
To convert the simulated reflectance spectrum into RGB values, we integrate the reflectance over the visible wavelength range using the CIE 1931 color-matching functions $S(\lambda)$ and a standard illuminant spectrum $I(\lambda)$, as shown in \Cref{eq:rgb_integral}:

\begin{equation}
x = \int_{\lambda_\text{min}}^{\lambda_\text{max}}
S(\lambda)I(\lambda)R(\lambda)d\lambda
\approx
\sum_{i=1}^{D} S(\lambda_i) I(\lambda_i)R(\lambda_i),
\label{eq:rgb_integral}
\end{equation}

\noindent where $S(\lambda)\in\mathbb{R}^{3\times1}$ represents the three color-matching functions used to obtain the $(R,G,B)$ components. In matrix form, the conversion can be expressed as
$x=S^\top(I\circ R)$,
where $\circ$ denotes element-wise multiplication.

\noindent\textbf{Implementation.}
We implement the multilayer optical model using the Python package \texttt{tmm}~\cite{byrnes2020multilayeropticalcalculations}. For each material, including graphene, MoS$_2$, WSe$_2$, and h-BN, we define the wavelength-dependent refractive index functions $n(\lambda)$ and $k(\lambda)$. We then compute the reflectance spectrum for each material and layer thickness. We convert the resulting reflectance spectra into RGB values and use them to render synthetic flakes that are combined with real microscopy backgrounds.

\subsection{Details of Synthia}

\textbf{Synthia} generates synthetic 2D material flakes across different materials, substrates, and layer thicknesses. Its core component is the multilayer optical model described in \Cref{sec:tmm}, which simulates thin-film interference to determine the expected color and contrast of each flake.

However, the optical model alone does not account for all of the variations observed in real microscopy images. The same material and layer thickness can appear different under changes in microscope illumination, camera white balance, and substrate color. To account for these differences, Synthia includes two additional components: the \textit{White-Balance-Aware} and \textit{Substrate-Aware} modules. The White-Balance-Aware module adjusts the generated colors according to the color characteristics of the target microscopy image, while the Substrate-Aware module identifies valid substrate regions for physically plausible flake placement.

\begin{figure}[!htbp]
\centering
\includegraphics[width=1\linewidth]{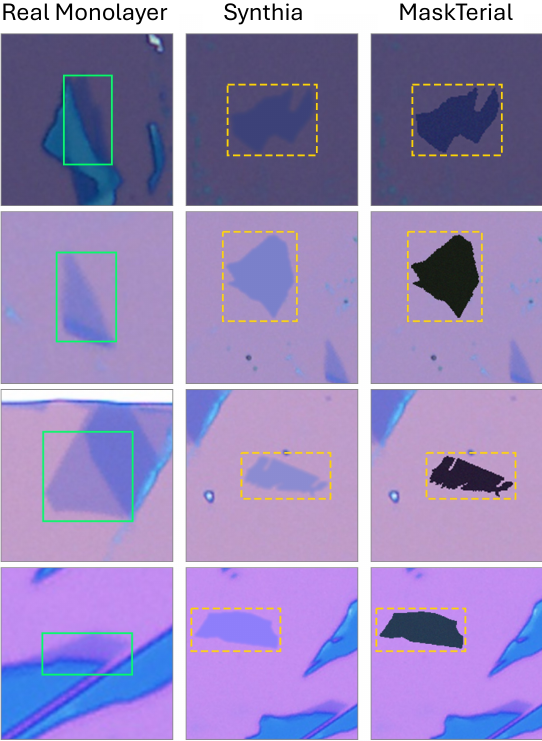}
\caption{\textbf{Qualitative comparison of synthetic monolayer flakes.}
Real monolayer flakes are shown alongside synthetic samples generated by Synthia and MaskTerial. Synthia produces subtler flake--substrate contrast that more closely resembles the real monolayer examples, while MaskTerial produces stronger visual separation from the surrounding substrate. \textbf{(Best viewed in color.)}}
\label{fig:optical_model_data_comparison}
\end{figure}

Synthia uses a Physics-Informed Attention (PIA) module, denoted by $\mathcal{F}_{\text{PIA}}$, to identify substrate regions that are suitable for synthetic flake placement. This prevents the generated flakes from being placed over existing flakes or other unsuitable image regions. The PIA module is described in \Cref{sec:pia}, and the complete Synthia generation procedure is provided in \Cref{alg:pgs}. \Cref{fig:optical_model_data_comparison} compares Synthia with MaskTerial and real monolayer flakes, illustrating how Synthia better preserves the subtle flake--substrate contrast observed in real microscopy images.
{
\vspace{-2mm}
\setlength{\textfloatsep}{0pt}
\begin{algorithm}[!b]
\caption{\textbf{Synthetic 2D Materials (Synthia)}}
\label{alg:pgs}
\footnotesize
\begin{algorithmic}[1]
\Require Reference microscopy image $\mathbf{I}_{\text{ref}}$, Optical model $\mathcal{T}$, 
Color projector $\Phi$, Material type $s$, Physics-Informed Attention function $\mathcal{F}_{\text{PIA}}$
\Ensure Synthetic microscopy image $\mathbf{I}_{\text{out}}$

\vspace{0.5em}
\Statex \textbf{// --- White-Balance-Aware Module ---}
\State $\mathbf{c}_{\text{sub}}^{\text{ref}} \gets \text{MedianColor}(\mathbf{I}_{\text{ref}})$
\State $R_{\text{sub}}(\lambda) \gets \mathcal{T}(\{\text{air}, \text{SiO}_2, \text{Si}\}, t_{\text{sub}}, \lambda)$
\State $\mathbf{c}_{\text{sub}}^{0} \gets \Phi(R_{\text{sub}}(\lambda), I(\lambda))$
\State $\mathbf{g} \gets \mathbf{c}_{\text{sub}}^{\text{ref}} \oslash \mathbf{c}_{\text{sub}}^{0}$ 
\Comment{Compute personalized white-balance gain based on reference substrate color}

\vspace{0.5em}
\Statex \textbf{// --- Substrate-Aware Module ---}
\State $\mathbf{A} \gets \mathcal{F}_{\text{PIA}}(\mathbf{I}_{\text{ref}})$
\State $\tilde{\mathbf{A}} \gets \text{Normalize}(\mathbf{A})$
\State $\mathbf{M}_{\text{sub}} \gets \mathbf{1}\{\tilde{\mathbf{A}} < \operatorname{Perc}_{90}(\tilde{\mathbf{A}})\}$ 
\Comment{Detect clean substrate areas for valid flake placement}

\vspace{0.5em}
\Statex \textbf{// --- Synthetic Flake Generation ---}
\For{$i = 1$ to $N_{\text{flakes}}$}
    \State $\mathbf{M}_i \gets \text{RandomFlakeMask()}$
    \State $t_i \gets \text{SampleThickness}(s)$
    \State Find valid coordinates $(u_i, v_i)$ such that 
           $\mathbf{M}_{\text{sub}}[v_i:v_i+h_i,\,u_i:u_i+w_i] \odot \mathbf{M}_i \equiv 0$
           \Comment{Avoid overlapping flakes}
    \State $R_i(\lambda) \gets \mathcal{T}(\{\text{air}, s, \text{SiO}_2, \text{Si}\}, t_i, \lambda)$
    \State $\mathbf{c}_i \gets \Phi(R_i(\lambda), I(\lambda))$ \Comment{Compute flake color via optical model}
    \State $\mathbf{c}_i \gets \mathbf{g} \odot \mathbf{c}_i$ 
           \Comment{Apply White-Balance-Aware correction}
    \State $\mathbf{I}_{\text{ref}}[v_i:v_i+h_i,\,u_i:u_i+w_i] 
           \gets \mathbf{c}_i \mathbf{M}_i 
           + \mathbf{I}_{\text{ref}}[v_i:v_i+h_i,\,u_i:u_i+w_i](1 - \mathbf{M}_i)$
\EndFor

\vspace{0.5em}
\State $\mathbf{I}_{\text{out}} \gets \mathbf{I}_{\text{ref}}$
\State \textbf{Return:} $\mathbf{I}_{\text{out}}$

\end{algorithmic}
\end{algorithm}
}

\subsection{QMat-Instruct Dataset}
\label{sec:qmat_instruct}

\noindent\textbf{Lack of an Instruction Dataset for MLLMs.}
Previous work in 2D material characterization has mainly formulated the problem as image-based detection, classification, or segmentation. These datasets provide labels such as bounding boxes, masks, or layer categories, but they are not designed to train MLLMs to interpret microscopy images and produce structured responses about the flakes. To our knowledge, no large-scale multimodal instruction dataset is currently designed for quantum flake characterization.

\noindent\textbf{Proposed QMat-Instruct Dataset.}
To support multimodal instruction tuning in this domain, we construct \textbf{QMat-Instruct}, a large-scale instruction dataset for quantum material characterization. Synthia allows us to generate microscopy images containing flakes with controlled material types, layer numbers, locations, and optical properties. These images provide the visual inputs and verified annotations needed to generate instruction--response pairs at scale.

Each QMat-Instruct sample contains a microscopy image, an instruction, and a structured response. The instructions cover common tasks in quantum material characterization, including flake counting, localization, layer identification, and binary verification. Example instructions include: "How many monolayer flakes are in the image?", "Locate the monolayer flakes" and "Does this sample contain a monolayer flake?"

\begin{figure*}[!htbp]
\centering
\includegraphics[width=0.8\linewidth]{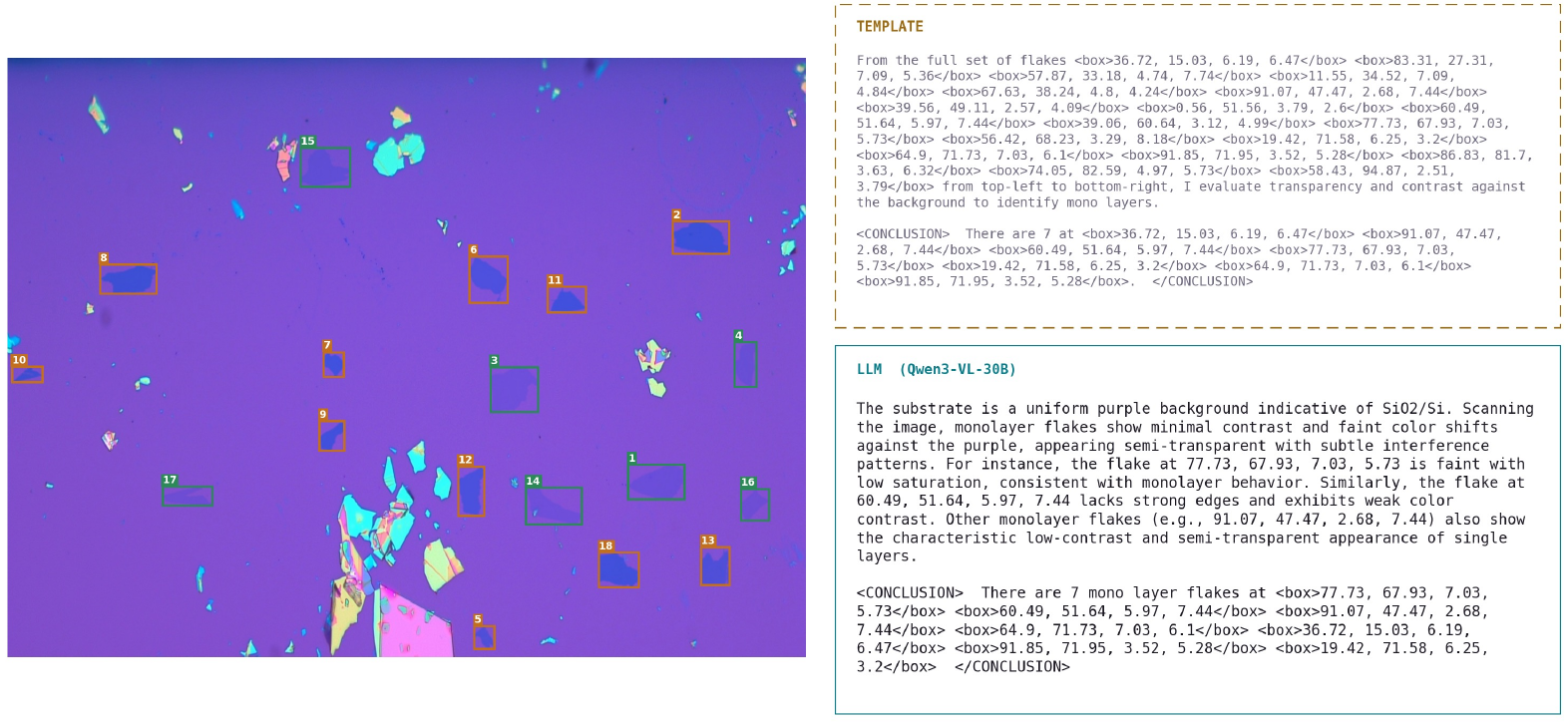}
\caption{\textbf{Comparison between template-based and LLM-generated reasoning for monolayer identification.}
Given the same microscopy image and candidate flake regions, the template-based approach follows a fixed, predefined reasoning structure to enumerate candidates and determine the final monolayer predictions. In contrast, the LLM generates a free-form, image-grounded explanation that describes the optical appearance of individual candidates, including contrast, color shift, transparency, and interference cues, before producing the final prediction.}
\label{fig:llm_vs_cot}
\end{figure*}

In the conference version of QMat-Instruct, we generated reasoning responses using predefined templates based on the corresponding flake annotations. While these templates provide structured physics-aware supervision, they do not account for the specific optical appearance of each image. In this journal extension, we instead generate an image-specific reasoning trace for each sample using a vision--language model together with the verified ground-truth annotations. \Cref{fig:llm_vs_cot} showcases the differences in how the VLM generates image--grounded reasoning rather than injecting a templated explanation. 

The generated reasoning is constrained to evidence that is visible in the optical microscopy image. This includes the relative contrast, color, boundaries, and appearance of each flake with respect to the surrounding substrate. The reasoning is also restricted from using information that cannot be obtained from the input image, such as AFM, Raman spectroscopy, photoluminescence, or other external measurements. Rather than relying on fixed absolute colors, the generated descriptions emphasize relative optical cues so that the supervision can be applied across different materials, substrates, and imaging conditions.

As a result, QMat-Instruct provides both structured task supervision and image-specific reasoning supervision for quantum flake characterization. The dataset teaches the MLLM not only to identify and localize flakes, but also to connect their visible optical appearance with the predicted layer category through both vision and language.

\section{The Proposed Method}

\begin{figure*}[!t]
\centering
\includegraphics[width=0.9\linewidth]{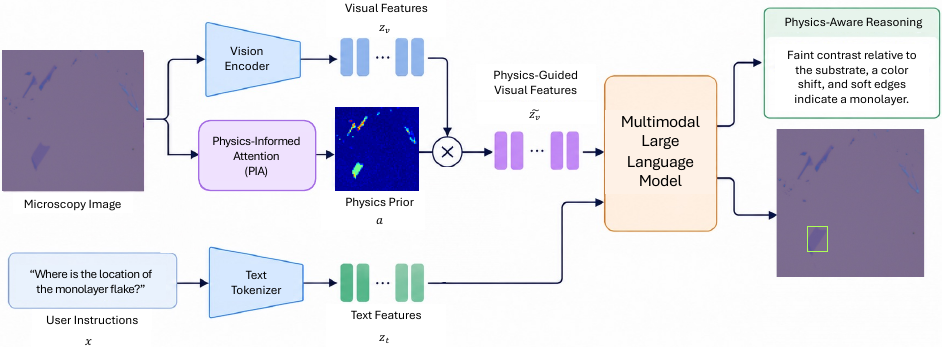}
\caption{Overview of the proposed \METHOD{} framework. Physics-Informed Attention (PIA) extracts optical contrast priors from the microscopy image to modulate the visual features, producing physics-guided representations that are combined with the text instruction in the MLLM for physics-aware reasoning and flake localization.}
\label{fig:architecture}
\end{figure*}

\METHOD\ takes an optical microscopy image $\mathbf{I}\in\mathbb{R}^{H\times W\times C}$ together with a textual instruction $\mathbf{T}=\{T_1,\ldots,T_L\}$ and generates a response that characterizes the visible 2D material flakes. As illustrated in \Cref{fig:architecture}, the output can include the number of detected flakes, their spatial locations, and their predicted layer categories.

The framework contains three main components: \textbf{(1)} a Vision Transformer (ViT) for extracting visual features from the microscopy image, \textbf{(2)} a Physics-Informed Attention (PIA) module for providing an optical contrast prior, and \textbf{(3)} an MLLM that combines the image and instruction representations to generate the final response.

\subsection{Vision Encoder with Physics-Informed Attention (PIA)}
Given the microscopy image $\mathbf{I}$, the vision encoder $\texttt{Enc}_v$ produces a sequence of $k$ patch-level visual representations as defined in \Cref{eq:vision-encoder}:

\begin{equation}
\label{eq:vision-encoder}
\begin{aligned}
z_v &= \texttt{Enc}_v(\mathbf{I}), \\
z_v &= [\,\mathbf{v}_1,\dots,\mathbf{v}_k\,],
\qquad
\mathbf{v}_i \in \mathbb{R}^{d}.
\end{aligned}
\end{equation}

To incorporate the optical appearance of the sample into these representations, PIA provides one contrast value for each corresponding image patch. As detailed in \Cref{sec:pia}, these values are collected as $\boldsymbol{\alpha}=[\alpha_1,\dots,\alpha_k]\in[0,1]^k$ and quantify the difference between each patch and the surrounding substrate in CIELAB space. We allow this prior to be calibrated during training because the measured appearance may vary with factors such as illumination, white balance, camera response, and substrate appearance.

\subsection{Learnable PIA Correction}
The PIA scores provide a physical prior, but they are computed directly from the image and cannot adapt during training. We therefore apply a small learnable correction to the PIA values. For each patch $i$, the corrected attention weight is represented as illustrated in \Cref{eq:weights_correction}:

\begin{equation}
\label{eq:weights_correction}
\beta_i = \sigma(w\alpha_i+b),
\qquad
\beta_i \in [0,1],
\qquad
w,b \in \mathbb{R},
\end{equation}

where $\sigma(\cdot)$ is the logistic sigmoid function. The learnable parameters $w$ and $b$ allow the model to adjust the scale and bias of the PIA scores during training without requiring an additional supervision signal.

Then, \Cref{eq:visual_token_corrected}, shows how the corrected weight is then applied to the corresponding visual token:
\begin{equation}
\label{eq:visual_token_corrected}
\tilde{\mathbf{v}}_i
=
\beta_i\mathbf{v}_i,
\qquad
\tilde{z}_v
=
[\,\tilde{\mathbf{v}}_1,\dots,\tilde{\mathbf{v}}_k\,].
\end{equation}

This allows the model to adjust the PIA scores during training while still using the optical contrast as a physical prior.

\subsection{Text Encoder}
In parallel with the visual pathway, the $L$ instruction tokens are mapped to their corresponding textual representations by $\texttt{Enc}_t$, as defined in \Cref{eq:text-encoder}:

\begin{equation}
\label{eq:text-encoder}
\begin{aligned}
z_t &= \texttt{Enc}_t(\mathbf{T}), \\
z_t &= [\,\mathbf{t}_1,\dots,\mathbf{t}_L\,],
\qquad
\mathbf{t}_j \in \mathbb{R}^{d}.
\end{aligned}
\end{equation}

The instructions in QMat-Instruct cover tasks such as flake counting, localization, layer identification, and verification. The corresponding responses provide supervision about the predicted flakes and the optical properties used to distinguish their layer categories.

\subsection{Multimodal Fusion and Decoding}
After applying the physics-informed modulation, the resulting visual sequence $\tilde{z}_v$ is paired with the encoded instruction $z_t$ to condition the MLLM generation process. The resulting output sequence is defined in \Cref{eq:llm}:

\begin{equation}
\label{eq:llm}
\mathcal{Y}
=
\texttt{LLM}(\tilde{z}_v,z_t),
\qquad
\mathcal{Y}
=
[\,y_1,\dots,y_M\,],
\end{equation}

where $\mathcal{Y}$ is the generated response and $M$ is the number of output tokens. Depending on the instruction, the response can contain the number of flakes, their locations, predicted layer categories, and the optical evidence used to support the predictions.

\subsection{Training Objective}
We train \METHOD\ using the standard autoregressive language-modeling loss over the QMat-Instruct responses as shown in \Cref{eq:loss-lm}:

\begin{equation}
\label{eq:loss-lm}
\mathcal{L}_{\text{LM}}
=
-\frac{1}{M}
\sum_{j=1}^{M}
\log
P_\theta
\left(
y_j
\mid
y_{<j},
\tilde{z}_v,
z_t
\right).
\end{equation}

No separate supervision is applied to the PIA correction. The parameters $(w,b)$ are optimized together with the model through the language-modeling loss. During inference, given a microscopy image and instruction $(\mathbf{I},\mathbf{T})$, the model generates the requested flake characterization.

\subsection{Physics-Informed Attention (PIA)}
\label{sec:pia}
\noindent\textbf{Motivation.}
Physics-Informed Attention (PIA) provides the vision encoder with an image-derived prior that indicates regions with optical differences from the surrounding substrate. These differences are governed by thin-film interference between the material, oxide layer, and underlying substrate, resulting in layer-dependent changes in observed color and intensity.

This behavior can be modeled using the transfer-matrix formulation described in ~\Cref{sec:tmm}. Directly applying this model requires acquisition-specific information such as wavelength-dependent optical constants, illumination characteristics, substrate parameters, and camera response, but these quantities are not always available for microscopy images collected across different experimental settings.

Thus, we estimate the relevant optical contrast from the observed image. By using the PIA, we compare the appearance of each image region with the surrounding substrate in CIELAB space, where the luminance $L^*$ and chromatic information $a^*$ and $b^*$ are represented separately. The resulting color difference provides an image-space proxy for the optical variations associated with the presence and thickness of a 2D material.

\subsection{PIA Approximation of Optical Contrast}

To construct the PIA prior from the microscopy image, we measure how strongly the appearance of each spatial region differs from the estimated substrate. First, we relate this observed difference to the image-formation process. Let $R(x,\lambda)$ denote the spectral reflectance at position $x$, and let $R_{\text{bg}}(\lambda)$ denote the reflectance of the bare substrate. For a camera channel $c$, the measured intensity depends on the reflectance, illumination spectrum $E(\lambda)$, and camera sensitivity $S_c(\lambda)$ as shown in ~\Cref{eq:pia_rgb}:

\begin{equation}
\label{eq:pia_rgb}
\begin{aligned}
I_c(x)
&=
\int
R(x,\lambda)
E(\lambda)
S_c(\lambda)
\,d\lambda, \\
I_c^{\text{bg}}
&=
\int
R_{\text{bg}}(\lambda)
E(\lambda)
S_c(\lambda)
\,d\lambda.
\end{aligned}
\end{equation}

The corresponding measurements across the three color channels form
$\mathbf{I}(x)=(I_R,I_G,I_B)$ and
$\mathbf{I}_{\text{bg}}=(I_R^{\text{bg}},I_G^{\text{bg}},I_B^{\text{bg}})$ for
the substrate. The RGB values are converted to CIELAB space using the mapping in \Cref{eq:pia_cielab}:

\begin{equation}
\label{eq:pia_cielab}
\Phi:\mathbb{R}^{3}\rightarrow\mathbb{R}^{3},
\qquad
(L^*,a^*,b^*)=\Phi(\mathbf{I}).
\end{equation}

Then, the PIA quantifies the optical difference at position $x$ using the Euclidean distance between the CIELAB representation of the observed region and the substrate as shown in \Cref{eqn:PIA-att}:

\begin{equation}
\label{eqn:PIA-att}
\begin{aligned}
\Delta E(x)
&=
\left\|
\Phi(\mathbf{I}(x))
-
\Phi(\mathbf{I}_{\text{bg}})
\right\|_2 \\
&=
\sqrt{
(L^*(x)-L^*_{\text{bg}})^2
+
(a^*(x)-a^*_{\text{bg}})^2
+
(b^*(x)-b^*_{\text{bg}})^2
}\\.
\end{aligned}
\end{equation}

A larger $\Delta E(x)$ indicates a stronger deviation from the substrate appearance and assigns greater importance to that region in the PIA prior. A derivation of \Cref{eqn:PIA-att} is provided in the appendix. Since an explicit substrate mask is not assumed, we estimate $\mathbf{I}_{\text{bg}}$ from the dominant background appearance of the microscopy image and evaluate the contrast over non-overlapping patches to construct the PIA map.

\subsection{Implementation}
We compute the PIA prior at the patch resolution used by the vision encoder. For each microscopy image, the substrate appearance is approximated by the median RGB value over the image, while each patch $\{\mathcal{P}_k\}$ is represented by its median RGB value as represented by \Cref{eq:pia_implementation}:

\begin{equation}
\label{eq:pia_implementation}
\begin{aligned}
\mathbf{I}_{\text{bg}}
&=
\operatorname{median}_{x\in\text{image}}
\mathbf{I}(x), \\
\widehat{\mathbf{I}}_k
&=
\operatorname{median}_{x\in\mathcal{P}_k}
\mathbf{I}(x).
\end{aligned}
\end{equation}

Using these estimates, we assign each patch a contrast value based on its CIELAB distance from the estimated substrate, as shown in \Cref{eq:pia_score}:

\begin{equation}
\label{eq:pia_score}
\text{PIA-score}[k]
=
\left\|
\Phi(\widehat{\mathbf{I}}_k)
-
\Phi(\mathbf{I}_{\text{bg}})
\right\|_2.
\end{equation}

\begin{figure}[!b]
\centering
\includegraphics[width=0.99\linewidth]{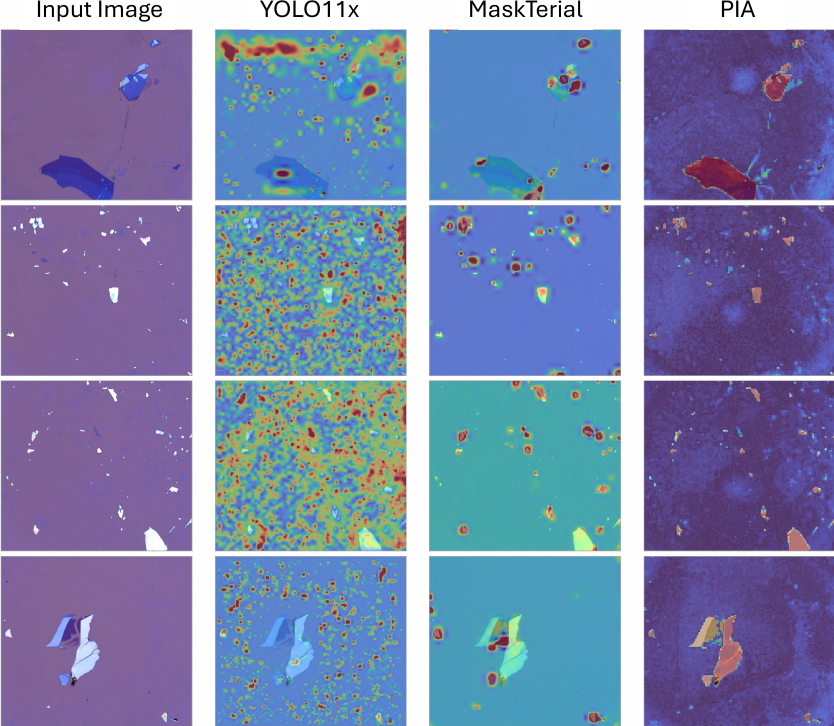}
\caption{Comparison of Grad-CAM for YOLO11x, query cross-attention for MaskTerial, and the proposed physics-informed attention signal $\alpha$. YOLO11x responds broadly to substrate texture, and MaskTerial mainly highlights regions after committing to an object. In contrast, $\alpha$ is derived from substrate-relative optical contrast and localizes flake regions without labels or retraining across materials and microscopes. \textbf{(Best viewed in color.)}}
\label{fig:pia_map}
\end{figure}

The collection of patch-level scores forms the PIA map. Higher values emphasize regions whose optical appearance differs more strongly from the substrate, providing the model with a spatial prior derived from the microscopy image without requiring material-specific optical parameters. \Cref{alg:lab_attention} summarizes the computation, while \Cref{fig:pia_map} compares the PIA signal $\alpha$ with Grad-CAM responses from YOLO11x and query cross-attention from MaskTerial. Across the examples, $\alpha$ more directly localizes substrate-relative optical differences around flake regions while suppressing diffuse responses over the surrounding substrate.

{
\begin{algorithm}[t]
\caption{\textbf{Physics-Informed Attention Module}}
\label{alg:lab_attention}
\footnotesize
\begin{algorithmic}[1]
\Require Image $\mathbf{I} \in \mathbb{R}^{H \times W \times 3}$,
image size $(H, W)$, patch size $(h, w)$
\Ensure Perceptual attention map $\mathbf{A}_{\text{LAB}} \in \mathbb{R}^{H \times W}$

\Function{PIA}{$\mathbf{I}, (H, W), (h, w)$}
    \State Resize $\mathbf{I}$ to $(H, W)$
    \State $\mathbf{I}_{\text{bg}} \gets \operatorname{median}_{x \in \mathbf{I}}(\mathbf{I}(x))$ %
    \State $\mathbf{I}_{\text{bg}}^{\text{LAB}} \gets \Phi(\mathbf{I}_{\text{bg}})$ %

    \State Initialize attention list $\mathcal{A} = [\,]$
    \For{each patch $\mathcal{P}_k$ in $\mathbf{I}$ of size $(h, w)$}
        \State $\widehat{\mathbf{I}}_k \gets \operatorname{median}_{x \in \mathcal{P}_k}(\mathbf{I}(x))$ %
        \State $\widehat{\mathbf{I}}_k^{\text{LAB}} \gets \Phi(\widehat{\mathbf{I}}_k)$ %
        \State $\Delta E_k = \big\| \widehat{\mathbf{I}}_k^{\text{LAB}} - \mathbf{I}_{\text{bg}}^{\text{LAB}} \big\|_2$
        \State Append $\Delta E_k$ to $\mathcal{A}$
    \EndFor

    \State Reshape $\mathcal{A}$ into $\mathbf{A}_{\text{LAB}}$ of size $(H, W)$
    \State Normalize $\mathbf{A}_{\text{LAB}} \gets (\mathbf{A}_{\text{LAB}} - \min)/(\max - \min)$
    \State \Return $\mathbf{A}_{\text{LAB}}$
\EndFunction
\end{algorithmic}
\end{algorithm}
}

\section{Image-Grounded Reasoning Supervision}
\label{sec:llmcot}

\subsection{Image-Specific Reasoning}

The conference version of \METHOD{} used predefined reasoning templates to construct the supervision for each training example. Verified flake locations and layer categories were inserted into these templates to produce a consistent response structure. While this provided stable supervision for the expected output format, the resulting explanations were largely shared across samples and did not account for the optical appearance of an individual microscopy image.

In this work, we instead generate a reasoning trace that is specific to each training image. An annotator vision--language model (VLM) receives the microscopy image, along with the verified flake locations and layer labels, and is asked to describe the visible optical evidence that supports the provided annotations. As a result, different samples with the same layer category can receive different explanations based on their observed contrast, color, boundaries, and appearance relative to the surrounding substrate.

\subsection{Reasoning Generation from Verified Ground Truth}

Synthia provides the exact locations and layer labels of the flakes inserted into each synthetic microscopy image. We therefore separate reasoning generation from detection and classification. Rather than asking the annotator VLM to independently determine where the flakes are or which layer category they belong to, we provide these verified annotations directly and ask it only to explain the corresponding visual evidence.

The bounding boxes and layer labels are consequently fixed by the Synthia ground truth, while only the reasoning span is generated by the annotator. This prevents errors in annotator localization or classification from changing the supervision targets and keeps the generated explanation grounded in known labels.

The final response retains the same \texttt{\textless CONCLUSION\textgreater} structure used during instruction tuning and inference. The enumeration and conclusion are constructed from the verified annotations, while the template-based explanation is replaced by the image-specific reasoning generated by the VLM. This preserves a consistent output format while introducing sample-dependent supervision.

\subsection{Constraints on Generated Reasoning}

We constrain the annotator to generate explanations using only information that is available from the optical microscopy image. The full system prompt is given in ~\Cref{fig:annotator_prompt}.

\noindent\textbf{Visible Optical Evidence.}
The reasoning is restricted to directly observable properties of the flake and its surrounding substrate. These cues include relative contrast and brightness, thin-film interference color, transparency, edge visibility, and the uniformity of the flake interior.

We explicitly exclude evidence from external characterization techniques such as photoluminescence, Raman spectroscopy, second-harmonic generation, and Atomic Force Microscopy (AFM). Such measurements are not contained in the input image and are therefore unavailable to \METHOD{} at inference time. Restricting the supervision to visible optical evidence ensures that the generated reasoning is based on the same source of information available to the downstream model.

\noindent\textbf{Relative Optical Cues.}
The observed appearance of a 2D material can vary substantially with material type, oxide thickness, illumination, camera response, and other acquisition conditions. Because material identity and oxide thickness are not provided to \METHOD{} during inference, the annotator is not instructed to associate a layer category with a fixed absolute color or intensity.

Instead, each flake is described relative to the surrounding substrate and other visible regions in the image. For example, the annotator may describe a candidate as faint, low-contrast, brighter, darker, or more transparent than the local substrate. This formulation reduces dependence on acquisition-specific absolute appearance and more closely matches the information available during inference. This constraint concerns the construction of the reasoning supervision rather than an assumption of improved transfer. We evaluate the cross-domain behavior of both supervision strategies in Sec.~\ref{sec:crossdomain}, where their transfer performance is comparable.

\subsection{Annotator Model}
\label{sec:annotator}

We generate all reasoning traces using an open-source annotator, Qwen3-VL-30B-A3B-Thinking, served locally with vLLM across four GPUs. Before annotation, we resize each training image so its maximum side is $1024$ pixels. The annotator is instructed to produce a concise reasoning trace of two to four sentences, with generation limited to $512$ tokens using temperature $0.7$ and top-$p$ $0.95$.

Because the verified enumeration and conclusion are constructed separately from the ground-truth annotations, we retain only the reasoning portion of the annotator output. We remove any preamble, bounding boxes, or conclusion generated by the VLM during post-processing. The resulting reasoning span is then combined with the deterministic enumeration and \texttt{\textless CONCLUSION\textgreater} spans used by the instruction-tuning format.

We use an open-source annotator so that reasoning generation can be performed locally without dependence on an external API, and the resulting supervision can be released together with the dataset. The annotation pipeline is not tied to a particular VLM. Another annotator can be substituted while retaining the same microscopy images, verified annotations, prompts, and output constraints.

\begin{figure}[t]
\centering
\footnotesize
\begin{tabular}{@{}p{0.95\columnwidth}@{}}
\toprule
\ttfamily You are an expert in optical microscopy of 2D materials on an SiO2/Si
substrate. Given the image and the verified monolayer flakes, write 2 to 4
sentences of grounded visual analysis: first describe the substrate/background,
then explain why the monolayer flakes read as monolayers from their visible
optical contrast, interference color, and transparency RELATIVE to the substrate
and to the other (thicker) flakes in the image.
\smallskip

\ttfamily Rules: reason ONLY from what is visible; do NOT assume or state the
material or the oxide thickness; do NOT rely on absolute color; NEVER mention
Raman, AFM, photoluminescence, or any non-optical modality. Output ONLY the
analysis sentences -- no preamble, no `Okay'/`Let me', no lists, no bounding
boxes, no conclusion. \\
\bottomrule
\end{tabular}
\caption{System prompt used for reasoning generation. The two constraints of
Sec.~\ref{sec:llmcot} are enforced directly in the prompt: reasoning is restricted
to visible optical evidence, and flakes are described relative to the substrate
rather than by absolute color.}
\label{fig:annotator_prompt}
\end{figure}

\section{Experimental Results}
\label{sec:experiments}

We evaluate \METHOD{} from several perspectives. We first report the general and mono-layer flake detection results from the conference version. We then evaluate instruction following and spatial grounding, study the effect of the proposed image-grounded reasoning supervision, analyze the contribution of the physics-aware components, and evaluate generalization to a material that is absent from training.

\subsection{Implementation Details and Benchmarking}

\noindent\textbf{Implementation Details.}
Our framework uses InternViT as the visual backbone. We resize each microscopy image to $448\times448$ and divide it into non-overlapping $14\times14$ patches. The text encoder is initialized from Qwen3~\cite{yang2025qwen3}. We train the model with AdamW at a learning rate of $2\times10^{-5}$, weight decay of $1\times10^{-4}$, and a batch size of $4$. The model is trained for $2$ epochs using 16 A100 GPUs. We apply random cropping, horizontal flipping, and color jitter in CIELAB space to improve robustness to illumination and white-balance changes. During inference, we generate the responses using a decoding temperature of $T=0.2$.

\noindent\textbf{Training Dataset.}
We train \METHOD{} using a hybrid dataset containing both real microscopy images and physics-based synthetic samples. The real images contain flakes collected under different illumination and magnification settings on 180\,nm SiO$_2$/Si substrates, following experimental setups similar to prior work~\cite{masubuchi2020deep}. Each image contains approximately 30 annotated flakes.

To increase the number of scarce mono- and few-layer examples, we generate additional samples using physics-based thin-film simulation and color rendering that matches the appearance of real microscopy images. The training data covers BN, Graphene, MoS$_2$, MoSe$_2$, MoWSe$_2$, WS$_2$, WSe$_2$, and WTe$_2$. Each image is paired with multiple physics-aware instruction--response examples describing the visible flakes, their layer categories, and their optical properties. We generate 50{,}000 samples for each material, for a total of 400{,}000 training samples.

\noindent\textbf{Extension Setup.}
The controlled experiments introduced in this journal extension use a
separate configuration described in \Cref{sec:reasonmetrics}. We train the
compared models from scratch using the same setup so that they differ only
in the supervision target.
Compared with the configuration used for the released models, these experiments use a smaller training budget. Each model is trained for one epoch using four A100 GPUs with a per-device batch size of $1$ and gradient accumulation of $4$, resulting in an effective batch size of $16$. We use sequence packing with a maximum budget of $8192$ tokens per packed sample. The optimizer, learning rate, backbone, and decoding temperature ($T=0.2$) remain unchanged.

Every variant is trained using the same $10{,}000$-image subset. This corresponds to $2.5\%$ of the 400{,}000 samples used to train the released models in \Cref{tab:flake_identification}. These extension experiments therefore isolate the effect of the supervision signal under the same reduced training budget, rather than reproduce the absolute performance of the released models.

For AP evaluation, the confidence assigned to each predicted box is computed using the geometric mean token probability over the tokens corresponding to that box, weighted by the logarithm of the box area. We remove duplicate predictions using non-maximum suppression with an IoU threshold of $0.3$. We apply the same decoding and ranking procedure to every variant. Since the extension configuration and regenerated training corpus differ from those used for the released models in \Cref{tab:flake_identification}, we only compare absolute results between models within the same extension experiment.

\noindent\textbf{Extension Corpus.}
We regenerate the training corpus for this work using the same Synthia generation procedure. The real component consists of reviewed microscopy images from our in-house material collections acquired on 180\,nm SiO$_2$/Si substrates. We retain the reviewed flake annotations, allowing each generated composite to contain both real and synthetic flakes.

We generate synthetic mono- and few-layer flakes for seven of the eight benchmark materials: MoS$_2$, MoSe$_2$, WS$_2$, WSe$_2$, Graphene, hBN, and MoWSe$_2$. WTe$_2$ is deliberately excluded from the extension training corpus so that \Cref{sec:crossdomain} can evaluate transfer to a material that is never observed during training. The regenerated corpus contains 40{,}000 composite images, and each extension variant is trained on the same $10{,}000$-image subset.

\noindent\textbf{Train/Test Separation.}
No image from QF-Bench is used during training. WTe$_2$ is also excluded entirely from the extension training corpus. Although its optical constants are available, we do not use them during synthetic generation, and no real or synthetic WTe$_2$ flakes are observed during training. We use WTe$_2$ as the held-out material in \Cref{sec:crossdomain}.

\noindent\textbf{QF-Bench.}
We evaluate the models using \textbf{QF-Bench}, which combines public microscopy datasets with our in-house collection. The benchmark contains eight 2D materials imaged under different microscopy and substrate conditions. Each image contains bounding-box and layer annotations for mono-layer (1L), few-layer (2--4L), and thick (5+L) flakes. QF-Bench contains 280{,}526 annotated flakes in total. \Cref{tab:test-data-by-layer-mat} shows the distribution across the different materials and layer categories.

\begin{table}[!t]
\centering
\footnotesize
\setlength{\tabcolsep}{4.2pt}
\renewcommand{\arraystretch}{1.05}
\caption{Number of annotated flakes in QF-Bench for each material and layer category.}
\label{tab:test-data-by-layer-mat}
\begin{tabular}{lrrrr}
\toprule
\textbf{Material} & \textbf{Mono} & \textbf{Few} & \textbf{Thick} & \textbf{Total} \\
\midrule
BN        & 13      & 111     & 10{,}100  & 10{,}224  \\
Graphene  & 1{,}856 & 2{,}081 & 4{,}270   & 8{,}207   \\
MoS$_2$   & 246     & 536     & 108{,}352 & 109{,}134 \\
MoSe$_2$  & 24      & 32      & 1{,}327   & 1{,}383   \\
MoWSe$_2$ & 9       & 35      & 293       & 337       \\
WS$_2$    & 43      & 5       & 2{,}144   & 2{,}192   \\
WSe$_2$   & 207     & 35      & 6{,}195   & 6{,}437   \\
WTe$_2$   & 148     & 582     & 141{,}882 & 142{,}612 \\
\midrule
\textbf{Total}
& \textbf{2{,}546}
& \textbf{3{,}417}
& \textbf{274{,}563}
& \textbf{280{,}526} \\
\bottomrule
\end{tabular}
\end{table}

\subsection{Main Flake Detection Results}
\label{sec:main_results}

We first evaluate the released \METHOD{} models on general and mono-layer flake detection. \Cref{tab:flake_identification} compares \METHOD{} with standard object detectors and existing methods developed for quantum material characterization. We report COCO-style AP, AP$^{50}$, and AP$^{75}$.

\begin{table*}[!t]
\centering
\footnotesize
\renewcommand{\arraystretch}{1.08}
\caption{Performance comparison on general and mono-layer flake detection tasks. %
}
\label{tab:flake_identification}

\begin{tabular*}{\textwidth}{@{\extracolsep{\fill}}lcccccccc}
\toprule
& \multicolumn{3}{c}{\textbf{General Flake Detection}}
& \multicolumn{3}{c}{\textbf{Mono-Layer Flake Detection}}
& \multicolumn{2}{c}{\textbf{Complexity}} \\
\cmidrule(lr){2-4}
\cmidrule(lr){5-7}
\cmidrule(lr){8-9}

\textbf{Detector}
& \textbf{AP}
& \textbf{AP$^{50}$}
& \textbf{AP$^{75}$}
& \textbf{AP}
& \textbf{AP$^{50}$}
& \textbf{AP$^{75}$}
& \textbf{\#Param}
& \textbf{FLOPs} \\
\midrule

MaskRCNN-R50~\cite{he2017mask}
& 18.7 & 36.5 & 17.8
& 9.8 & 28.9 & 9.0
& 44M & 134.4B \\

MaskRCNN-R101~\cite{he2017mask}
& 20.3 & 38.1 & 18.9
& 11.1 & 30.6 & 10.3
& 63M & 334.8B \\

\midrule

ViTDet (base)~\cite{li2022exploringplainvisiontransformer}
& 17.4 & 35.2 & 16.5
& 8.7 & 27.6 & 7.8
& 86M & 1.3T \\

ViTDet (large)~\cite{li2022exploringplainvisiontransformer}
& 18.9 & 37.0 & 17.3
& 9.9 & 29.8 & 8.5
& 300M & 4.1T \\

\midrule

YOLOv11-m~\cite{Jocher_Ultralytics_YOLO26_Unified_2026}
& 24.8 & 41.9 & 21.1
& 17.0 & 27.2 & 16.4
& 20M & 68.0B \\

YOLOv11-l
& 26.9 & 43.0 & 23.5
& 18.9 & 29.8 & 17.3
& 25M & 86.9B \\

YOLOv11-x
& 29.6 & 44.5 & 24.6
& 19.8 & 31.3 & 18.7
& 57M & 194.9B \\

\midrule

Co-DETR~\cite{zong2023detrs}
& 22.7 & 35.9 & 23.3
& 12.2 & 25.3 & 11.4
& 304M & 119.4B \\

RT-DETRv2~\cite{lv2024rt}
& 27.5 & 37.2 & 27.0
& 17.5 & 37.8 & 14.7
& 76M & 259.0B \\

\midrule

MaskTerial~\cite{uslu2025maskterial}
& 23.8 & 41.0 & 25.4
& 16.8 & 35.5 & 17.2
& 45M & 150.7B \\

$\varphi$-Adapt~\cite{phiadapt2025physics}
& 30.3 & 49.1 & 27.4
& 24.1 & 38.2 & 23.4
& 91M & 368.3B \\

\midrule

\METHOD-1B
& 36.9 & 50.7 & 38.6
& 28.0 & 42.9 & 29.5
& 1.1B & 1.5T \\

\METHOD-2B
& 38.7 & 53.6 & 41.2
& 30.2 & 45.5 & 32.0
& 2.3B & 3.4T \\

\METHOD-4B
& 42.4 & 58.2 & 46.8
& 34.1 & 50.2 & 38.0
& 4.7B & 7.9T \\

\textbf{\METHOD-8B}
& \textbf{45.6}
& \textbf{60.5}
& \textbf{47.9}
& \textbf{37.3}
& \textbf{52.8}
& \textbf{39.9}
& 8.5B & 14.5T \\

\bottomrule
\end{tabular*}
\end{table*}

\begin{figure*}[!t]
    \centering
    \includegraphics[width=\textwidth]{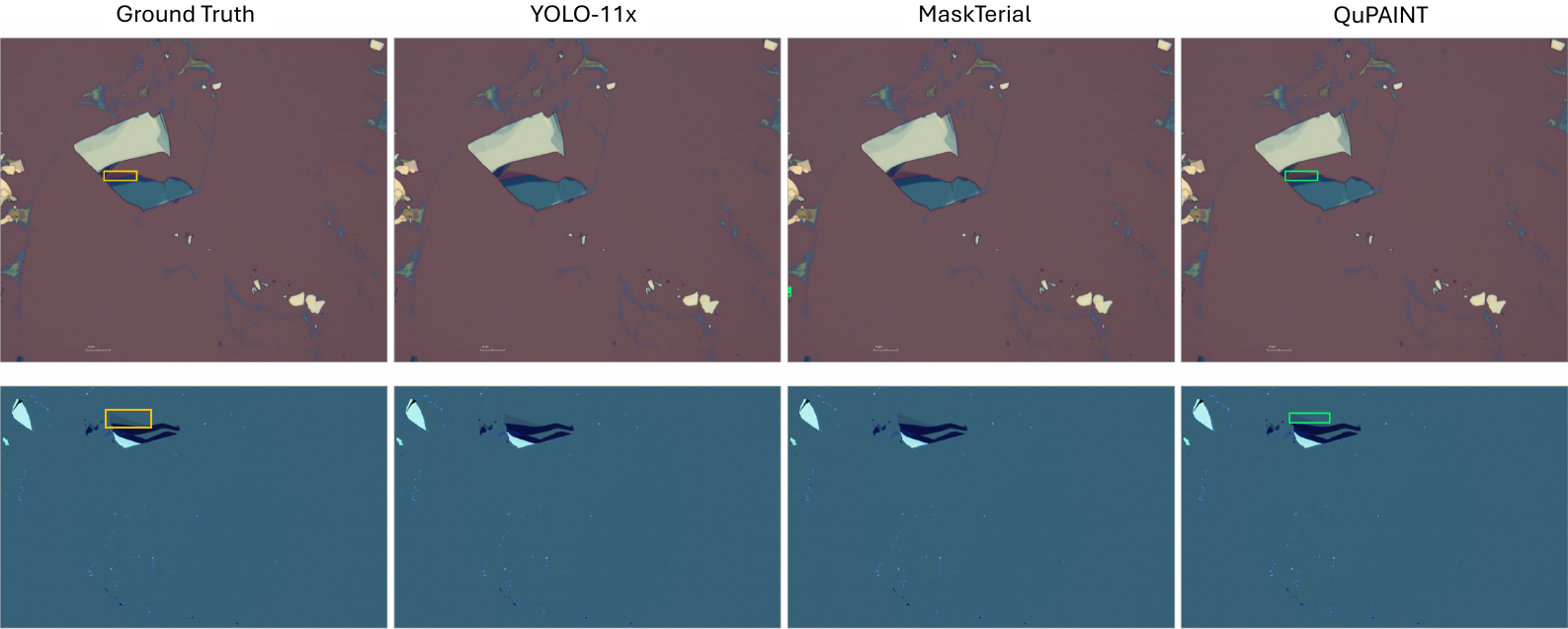}
    \caption{\textbf{Qualitative comparison of monolayer flake detection.}
Representative examples comparing ground-truth annotations with predictions from YOLO-11x, MaskTerial, and QuPAINT. QuPAINT more closely matches the ground-truth monolayer locations, while the baseline methods exhibit missed detections or spurious predictions. \textbf{(Best viewed in color.)}}\label{fig:qualitative_detection}
\end{figure*}

As shown in \Cref{tab:flake_identification}, \METHOD{} consistently improves as the model size increases and achieves the strongest performance across all reported detection metrics. \METHOD-8B reaches an AP of $45.6$ for general flake detection and $37.3$ for mono-layer detection. Mono-layer detection remains more difficult across all methods because these flakes generally exhibit weaker optical contrast with respect to the surrounding substrate. As shown in \Cref{tab:flake_identification}, \METHOD{} consistently improves as the model size increases and achieves the strongest performance across all reported detection metrics. \METHOD-8B reaches an AP of $45.6$ for general flake detection and $37.3$ for mono-layer detection. Mono-layer detection remains more difficult across all methods because these flakes generally exhibit weaker optical contrast with respect to the surrounding substrate. \Cref{fig:qualitative_detection} provides qualitative comparisons with YOLOv11-x and MaskTerial on QF-Bench. Across the examples, \METHOD-4B more closely matches the ground-truth mono-layer regions, while the baseline methods exhibit missed detections or spurious predictions.

\subsection{Instruction-Following and the Effect of Reasoning Supervision}
\label{sec:reasonmetrics}

To evaluate the two supervision strategies beyond standard detection AP, we report the instruction-following, grounding, and confidence metrics summarized in \Cref{tab:reasoning}. For counting, \textbf{Count Err.} denotes the mean absolute error between the predicted and ground-truth number of flakes. For localization, \textbf{Grnd.@50} and \textbf{Grnd.@75} denote grounding accuracy at IoU thresholds of $0.5$ and $0.75$, respectively, measuring the fraction of ground-truth targets successfully localized at each threshold. \textbf{Loc. Prec.@50} measures the fraction of predicted boxes that correspond to a ground-truth flake at IoU $0.5$. Finally, \textbf{Conf. AUC} denotes the area under the receiver operating characteristic curve obtained using the model's per-box confidence as the ranking score and prediction correctness as the binary target. Higher AUC indicates that the model more reliably assigns greater confidence to correct predictions than to incorrect ones.
We use this evaluation to isolate the main supervision change introduced in the journal extension. The conference version uses fixed reasoning templates, while the journal version replaces the template text with the image-specific reasoning generated from verified annotations described in \Cref{sec:llmcot}. The two models use the same backbone, $10{,}000$ training images, annotations, instruction phrasing, optimizer configuration, and decoding settings. The enumerated flake boxes are generated in a fixed top-left to bottom-right order, while the final conclusion is constructed deterministically from the ground-truth annotations for both models. Thus, the only difference between the two supervision targets is the reasoning span between the enumeration and final conclusion.

\begin{table}[!t]
\centering
\footnotesize
\setlength{\tabcolsep}{3.4pt}
\caption{Instruction-following, grounding, and confidence results on the $1{,}186$ QF-Bench images common to both runs. 
Confidence AUC is reported for the mono-layer task.}
\label{tab:reasoning}
\resizebox{\columnwidth}{!}{%
\begin{tabular}{lccccc}
\toprule
\textbf{Model}
& \textbf{Count Err.} $\downarrow$
& \textbf{Grnd.@50} $\uparrow$
& \textbf{Grnd.@75} $\uparrow$
& \textbf{Loc. Prec.@50} $\uparrow$
& \textbf{Conf. AUC} $\uparrow$ \\
\midrule
Template-Based
& \textbf{2.57}
& 29.3
& 13.8
& 18.3
& 0.820 \\
Image-Grounded
& 2.77
& 29.3
& \textbf{15.5}
& \textbf{19.6}
& \textbf{0.846} \\
\bottomrule
\end{tabular}%
}
\end{table}

Image-grounded reasoning does not increase the number of flakes the model successfully grounds. As shown in \Cref{tab:reasoning}, grounding accuracy at IoU $0.5$ is identical at $29.3$ for both supervision strategies. Thus, replacing the fixed templates with image-specific reasoning does not cause the model to find flakes that the template-based model would otherwise miss.

As further shown in \Cref{tab:reasoning}, the difference becomes more visible under
stricter localization and ranking criteria. Grounding accuracy at IoU $0.75$
increases from $13.8$ to $15.5$, while localization precision at IoU $0.5$
increases from $18.3$ to $19.6$.

For monolayer predictions matched to the ground truth, the mean IoU changes only from $0.6875$ to $0.6931$ for mono-layer flakes. General-flake mean IoU similarly changes from $0.7178$ to $0.7122$. These differences are small, indicating that the improvement is not primarily due to better-shaped or more tightly aligned boxes. Instead, the results suggest that the model identifies and ranks correct predictions more reliably.

This effect is most clearly reflected in the model's confidence ranking. We measure the AUC between the model's own per-box confidence and the actual correctness of each prediction. For mono-layer detection, confidence AUC increases from $0.820$ with template-based supervision to $0.846$ with image-grounded supervision. The same trend is observed for general flake detection, where AUC increases from $0.821$ to $0.826$. Unlike the task-specific changes in localization metrics, confidence improves consistently across both settings. Image-grounded reasoning therefore does not substantially change which flakes are detected, but it yields a ranking signal that separates correct predictions from incorrect ones more reliably.

Counting performance moves slightly in the opposite direction. The absolute counting error increases from $2.57$ to $2.77$. Both models primarily over-count the number of flakes, with a mean signed error of $+2.37$ for template-based supervision and $+2.54$ for image-grounded supervision. Thus, the image-grounded reasoning does not improve counting accuracy under this evaluation.

\noindent\textbf{Image-Specific Descriptions.}
Although the two models obtain similar localization performance, the reasoning text they generate differs substantially. The template-based model produces the same description on $75.3\%$ of benchmark images, and only $19.4\%$ of its descriptions are distinct. In comparison, the image-grounded model produces a distinct description on $81.5\%$ of the images.

We further evaluate whether this variation corresponds to measurable optical properties of the input image. We group descriptions by the optical vocabulary they use and compare them with the measured $\Delta E$ of the annotated flakes. Images whose descriptions contain low-contrast terms such as \emph{faint}, \emph{minimal}, or \emph{subtle} have a median flake $\Delta E$ of $20.1$. In comparison, images whose descriptions contain higher-contrast terms such as \emph{distinct}, \emph{pronounced}, or \emph{sharp} have a median $\Delta E$ of $21.6$. This difference is statistically significant under a Mann--Whitney test ($p=0.03$, $n=1{,}186$).

The generated descriptions are also consistent with the measured optical differences between layer categories. Using the ground-truth annotations, mono-layer flakes have a median $\Delta E$ of $8.1$, compared with $20.8$ for thick flakes. Thus, descriptions that characterize mono-layer flakes as having weak or minimal contrast align with the optical measurements in the benchmark rather than following a generic description. These results show that image-grounded supervision produces descriptions that are more specific to the individual image and are consistent with its measured optical properties, even when the two supervision strategies remain comparable in localization performance.

\subsection{Ablation: Physics-Aware Description and PIA}
\label{sec:ablation}

We next study the individual contributions of the Physics-Aware Description (PAD) and Physics-Informed Attention (PIA) components. \Cref{tab:pir_pia_ablation} reports the performance obtained when each component is enabled independently and when both are used together for mono-layer flake detection.

\begin{table}[!t]
\centering
\footnotesize
\setlength{\tabcolsep}{8pt}
\caption{Ablation of Physics-Aware Description (PAD) and Physics-Informed Attention (PIA) on mono-layer flake detection.}
\label{tab:pir_pia_ablation}
\begin{tabular}{ccccc}
\toprule
\textbf{PAD}
& \textbf{PIA}
& \textbf{AP}
& \textbf{AP$^{50}$}
& \textbf{AP$^{75}$} \\
\midrule
\xmark & \xmark & 28.1 & 42.7 & 30.3 \\
\xmark & \cmark & 30.6 & 44.8 & 32.6 \\
\cmark & \xmark & 31.7 & 46.5 & 33.4 \\
\cmark & \cmark & \textbf{34.1} & \textbf{50.2} & \textbf{38.0} \\
\bottomrule
\end{tabular}
\end{table}

Both components improve mono-layer detection independently. As shown in \Cref{tab:pir_pia_ablation}, adding PIA increases AP from $28.1$ to $30.6$, while PAD increases AP to $31.7$. Using both components provides the strongest result, reaching an AP of $34.1$, AP$^{50}$ of $50.2$, and AP$^{75}$ of $38.0$.

The combined improvement of $+6.0$ AP is close to the sum of the individual gains from PIA ($+2.5$) and PAD ($+3.6$). This suggests that the physics-aware textual supervision and physics-informed visual prior provide largely complementary information to the model. The final row corresponds to the released \METHOD-4B model in \Cref{tab:flake_identification} and reproduces its mono-layer detection results.

\subsection{Cross-Material and Unseen-Material Generalization}\label{sec:crossdomain}

The experiments in this subsection use the reduced-budget extension models described in \Cref{sec:reasonmetrics}, so their absolute values are not directly comparable with the released models in \Cref{tab:flake_identification}.
The optical appearance of a flake can change with the material, substrate, microscope, and imaging conditions. We therefore evaluate \METHOD{} on WTe$_2$, the only benchmark material that is completely excluded from the extension training corpus. This allows us to measure transfer to a material whose optical response is never observed during training.
As shown in the cross-material detection results in \Cref{tab:crossdomain}, image-grounded supervision consistently improves detection sensitivity across materials. Recall increases on seven of the eight evaluated materials and on six of the seven materials observed during training. The largest improvement occurs for Graphene, where recall increases from $37.4$ to $51.3$, followed by hBN ($68.6$ to $74.6$) and WSe$_2$ ($14.2$ to $18.7$). MoS$_2$ is the only material for which recall decreases by $0.5$ points.

The per-material results in \Cref{tab:crossdomain} show that the benefit of image-grounded supervision is not confined to a particular dataset size. F1 improves on larger material subsets such as Graphene and WSe$_2$, as well as on smaller subsets such as MoSe$_2$ and WS$_2$. The largest improvement occurs on Graphene, where F1 increases from $37.5$ to $43.4$ across 8,207 annotated flakes. Overall, macro-F1 increases from $27.1$ to $28.6$, indicating improved aggregate detection performance across material-dependent variations in optical appearance.

The held-out WTe$_2$ evaluation directly tests a material completely excluded from the extension training corpus. Despite WTe$_2$ containing 142,612 annotated flakes, the largest material subset in QF-Bench, both supervision strategies achieve the same F1 score of $23.0$. Thus, image-grounded supervision improves aggregate cross-material performance while preserving detection performance on the unseen material.

\begin{table}[!t]
\centering
\footnotesize
\setlength{\tabcolsep}{5.5pt}
\renewcommand{\arraystretch}{1.08}
\caption{Cross-material detection performance at IoU $0.5$. Image-grounded supervision improves macro-F1 across QF-Bench while preserving performance on the held-out WTe$_2$ material.}
\label{tab:crossdomain}
\begin{tabular}{lccc}
\toprule
\textbf{Material}
& \textbf{\#Flakes}
& \multicolumn{2}{c}{\textbf{F1}} \\
\cmidrule(lr){3-4}
&
& \textbf{Template-based}
& \textbf{Image-grounded} \\
\midrule

Graphene
& 8,207
& 37.5
& \textbf{43.4} \\

hBN
& 10,224
& \textbf{42.6}
& 42.5 \\

MoS$_2$
& 109,134
& \textbf{18.8}
& 17.9 \\

MoSe$_2$
& 1,383
& 18.6
& \textbf{20.6} \\

MoWSe$_2$
& 337
& \textbf{39.4}
& 38.4 \\

WS$_2$
& 2,192
& 17.7
& \textbf{20.5} \\

WSe$_2$
& 6,437
& 19.0
& \textbf{22.8} \\

\midrule
WTe$_2$ (held out)
& 142,612
& 23.0
& 23.0 \\

\midrule
\textbf{Macro-F1}
& --
& 27.1
& \textbf{28.6} \\

\bottomrule
\end{tabular}
\end{table}

\section{Conclusion and Limitations}

\noindent\textbf{Conclusions.}
In this work, we presented a complete framework for automated characterization of 2D quantum materials from optical microscopy images. We introduced \textbf{Synthia}, a physics-based synthetic data generation framework for producing realistic microscopy images across different materials and layer thicknesses. Using these images and their verified annotations, we constructed \textbf{QMat-Instruct}, a large-scale multimodal instruction dataset for quantum flake characterization. We further introduced \METHOD{}, which combines multimodal instruction tuning with a Physics-Informed Attention (PIA) prior for identifying and characterizing flakes from optical microscopy images.

In this work, we replace the template-based chain-of-thought supervision used in the conference version with image-specific reasoning generated from verified flake annotations and the optical evidence visible in each microscopy image. This allows the reasoning supervision to describe the appearance of individual flakes instead of relying on fixed templates. We also extend the evaluation through QF-Bench to study instruction following, grounding, reasoning supervision, and generalization in addition to standard flake-detection performance.

\noindent\textbf{Limitations.}
Our approach still has several limitations. Synthetic images cannot capture every variation observed across microscopes, substrates, defects, and imaging conditions, so a domain gap can remain when applying the model to new settings. In addition, the optical appearance of a flake depends on factors such as material type, oxide thickness, illumination, and camera response, which can make layer identification ambiguous from a single optical image. Finally, QF-Bench remains highly imbalanced toward thick flakes, while verified monolayer and few-layer samples are much rarer.

\section{Data Availability}
The data supporting the findings of this work are available from the corresponding author upon reasonable request.

\bibliographystyle{IEEEtran}
\bibliography{main}
\begin{IEEEbiography}
[{\includegraphics[width=1in,height=1.25in,clip,keepaspectratio]{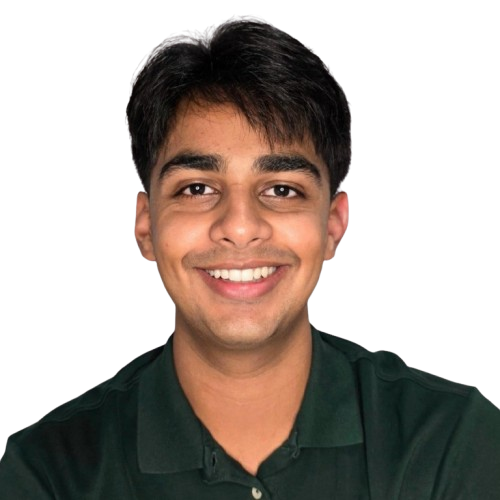}}]
{Sankalp Pandey} is a Ph.D. student in Computer Science and Distinguished Doctoral Fellow at the University of Arkansas. He works in the Quantum AI Lab and the Computer Vision \& Image Understanding Lab, where is advised by Professor Khoa Luu. He received B.S. degrees in Computer Science and Computer Engineering, with honors, from the University of Arkansas, along with minors in Mathematics and Data Analytics. His research focuses on multimodal and agentic learning systems for materials science. His work has been presented at venues including CVPR Findings, NeurIPS workshops, and IEEE Quantum Week workshops, and he contributes to the development of open-source deep learning tools for scientific discovery.
\end{IEEEbiography}

\begin{IEEEbiography}
[{\includegraphics[width=1in,height=1.25in,clip,keepaspectratio]{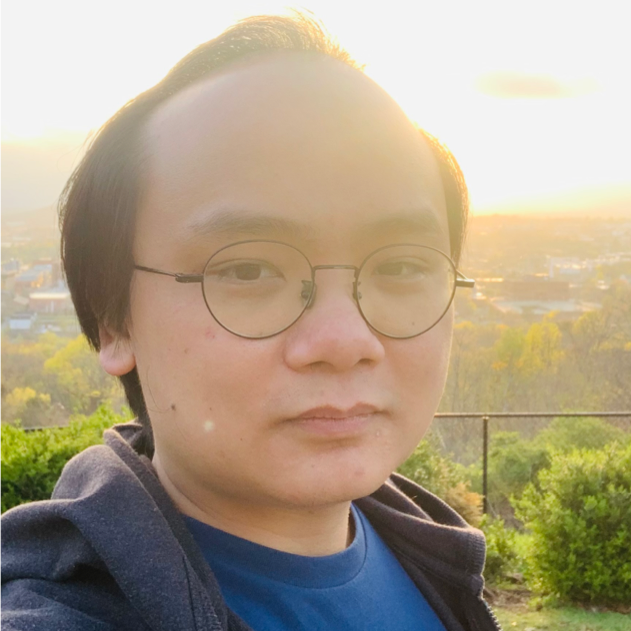}}]
{Xuan-Bac Nguyen} got PhD of CSCE - University of Arkansas in the United States, where he was under the guidance and supervision of Professor Khoa Luu. He holds a Master’s degree in Computer Science from Chonnam National University, where he worked alongside Professor Gueesang Lee. His academic journey began with a B.Sc. degree in Electronics and Telecommunication from the University of Engineering and Technology, Vietnam National University.
Bac’s research primarily focuses on areas such as Computer Vision, Unsupervised Learning, Self-supervised Learning, and more. He has consistently demonstrated his expertise in these fields and has actively contributed to the academic community. His research publications have been accepted in CVPR, TPAMI, and many others.
\end{IEEEbiography}

\begin{IEEEbiography}
[{\includegraphics[width=1in,height=1.25in,clip,keepaspectratio]{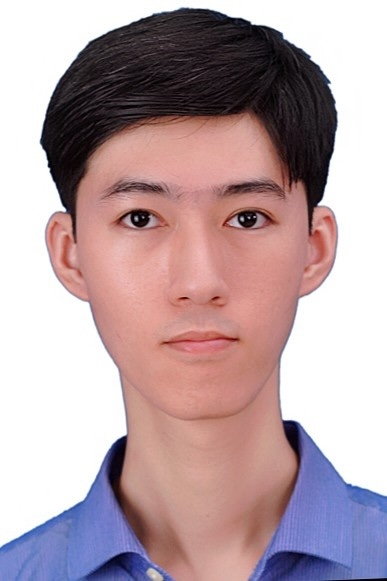}}]
{Hoang-Quan Nguyen} is a PhD candidate at the University of Arkansas, where he is supervised by Dr. Khoa Luu. He is a research assistant at the Computer Vision and Image Understanding Lab. In 2022, he received his B.Sc. degree in Computer Science from the Honors Program at the University of Science, VNU-HCM, under the supervision of Dr. Khoa Luu and Dr. Son Tran. His research interest includes computer vision, deep learning, and quantum machine learning.
\end{IEEEbiography}

\begin{IEEEbiography}
[{\includegraphics[width=1in,height=1.25in,clip,keepaspectratio]{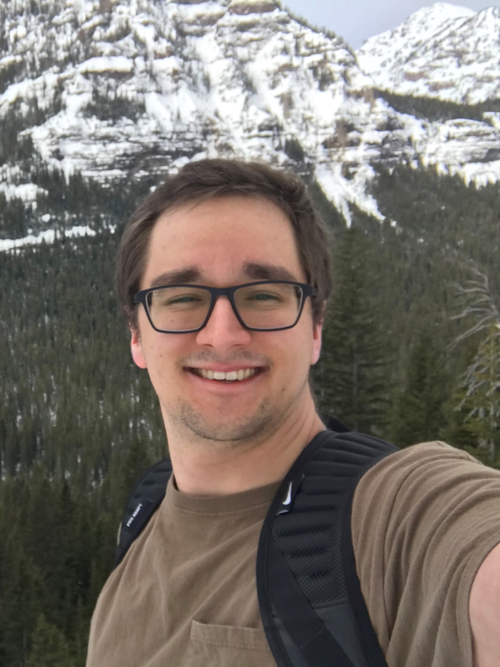}}]
{Tim Faltermeier}
is a PhD candidate in Physics at Montana State University working under the supervision of Dr. Nicholas Borys. He is a research assistant working on instrumentation development for the MonArk NSF Quantum Foundry. He received his B.Sc. in Physics from Wyoming State University and his M.Sc. in Physics from Montana State University. His research interests include the development of automated laboratory equipment and nano-optical spectroscopy of low-dimensional materials. 
\end{IEEEbiography}

\begin{IEEEbiography}
[{\includegraphics[width=1in,height=1.25in,clip,keepaspectratio]{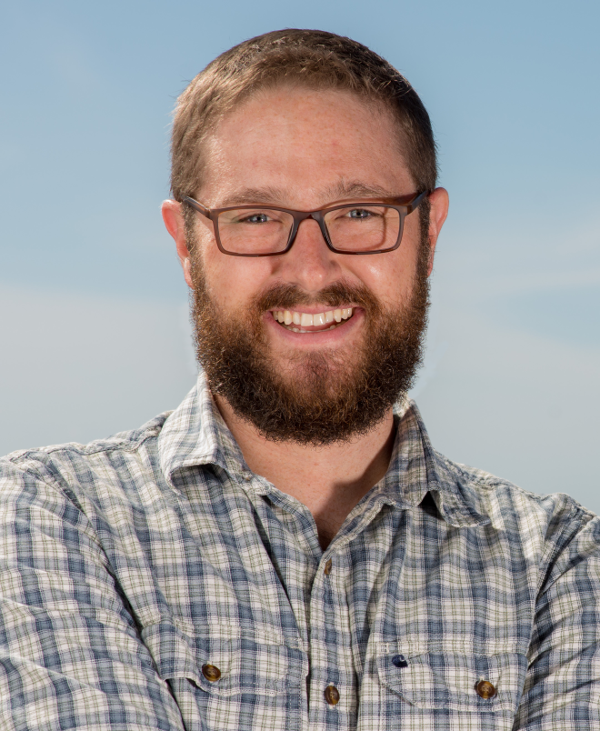}}]
{Nicholas Borys} is an Associate Professor of Physics at Montana State University and co‑associate Director of the MonArk NSF Quantum Foundry. He received his Ph.D. in Physics from the University of Utah, following earlier work in software development at Boeing and a B.S. in Mathematics and Computer Science from the Colorado School of Mines. His research focuses on quantum optics, 2D materials, and nano‑optical spectroscopy, with an emphasis on cryogenic and ultrafast characterization of low-dimensional quantum materials. Prior to joining MSU, he was a project scientist at the Molecular Foundry at Lawrence Berkeley National Laboratory. Dr. Borys currently leads an experimental research group advancing quantum materials and devices for emerging technologies in sensing, communication, and quantum information.
\end{IEEEbiography}

\begin{IEEEbiography}
[{\includegraphics[width=1in,height=1.25in,clip,keepaspectratio]{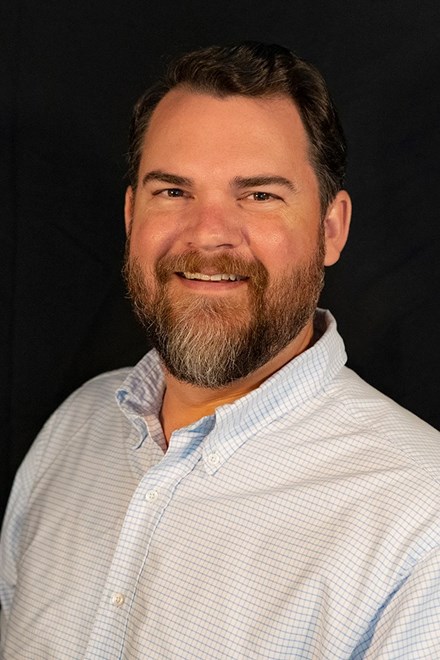}}]
{Hugh Churchill} received the Ph.D. degree in physics from Harvard University in 2012. He was a Pappalardo Postdoctoral Fellow in physics at MIT. He is currently a Professor and the 21st Century Chair of Nanophysics with the Department of Physics, University of Arkansas (UA). He is also the Associate Director of Operations at MonArk NSF Quantum Foundry. Since 2015, he has led the Quantum Device Laboratory, UA. His research interests include quantum materials and devices, low-dimensional materials, quantum transport, optoelectronics, and automation of experiments.
\end{IEEEbiography}

\begin{IEEEbiography}
[{\includegraphics[width=1in,height=1.25in,clip,keepaspectratio]{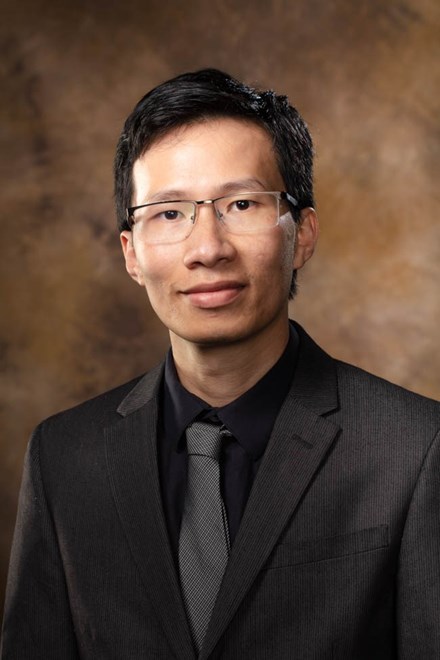}}]
{Khoa Luu} 
is an Associate Professor and the Director of the Quantum AI Lab and the Computer Vision and Image Understanding (CVIU) Lab in the Department of Computer Science \& Computer Engineering at the University of Arkansas, Fayetteville. He is affiliated with the NSF MonARK Quantum Foundry. 
He serves as an Associate Editor with ACM Computing Surveys and IEEE Transactions on Quantum Engineering. 
He is also an Area Chair for major AI conferences, including CVPR, NeurIPS, ICML, ICLR, WACV, and AAAI, and has organized the IEEE/CVF CVPR Precognition Workshop 2019-2026. He is the Program Chair of the International Conference on Quantum Computing Infrastructure and Technology Trento, Italy, 2026, Secure and Trustworthy Quantum Machine Learning (SaTQuML) at NeurIPS 2026, and IEEE GreenTech in 2024. Prior to joining the University of Arkansas, he was with CyLab Biometrics Center at Carnegie Mellon University (CMU), as Research Project Director. His research spans multimodal and generative AI, foundation models, trustworthy and continual learning, computer vision, biometrics, autonomous perception and human–robot intelligence, vision–brain modeling, quantum machine learning, and AI-assisted quantum-material discovery. He has coauthored more than 150 publications, eight patents and three Best Paper awards. He was selected to the National Academy of Inventors, University of Arkansas Chapter, 2024. He was a vice chair of the Montreal Chapter of IEEE SMCS in Canada from September 2009 to March 2011. 
\end{IEEEbiography}

\end{document}